\documentclass[pdflatex,sn-mathphys-num,iicol]{sn-jnl}

\usepackage{algorithm}
\usepackage{algorithmic}
\usepackage{graphicx}%
\usepackage{multirow}%
\usepackage{amsmath,amssymb,amsfonts}%
\usepackage{amsthm}%
\usepackage{mathrsfs}%
\usepackage[title]{appendix}%
\usepackage{xcolor}%
\usepackage{textcomp}%
\usepackage{manyfoot}%
\usepackage{booktabs}%
\usepackage{listings}%
\usepackage{gensymb}
\usepackage{makecell}
\usepackage{diagbox}
\usepackage{threeparttable}
\usepackage{tabularx}
\usepackage{float}

\theoremstyle{thmstyleone}%
\theoremstyle{thmstyletwo}%

\theoremstyle{thmstylethree}%

\begin{document}
	
	\title[Article Title]{FBSDiff++: Improved Frequency Band Substitution of Diffusion Features for Efficient and Highly Controllable Text-Driven Image-to-Image Translation}
	
	
	
	
	
	
	
	\author[1]{\fnm{Xiang} \sur{Gao}}\email{gaoxiang1102@bupt.edu.cn}
	\author*[1]{\fnm{Yunpeng} \sur{Jia}}\email{xibei156@bupt.edu.cn}
	\affil[1]{
		\orgname{Beijing University of Posts and Telecommunications}, \orgaddress{\city{Beijing}, \postcode{102206}, \country{China}}}
	
	
	
	\abstract{With large-scale text-to-image (T2I) diffusion models achieving significant advancements in open-domain image creation, increasing attention has been focused on their natural extension to the realm of text-driven image-to-image (I2I) translation, where a source image acts as visual guidance to the generated image in addition to the textual guidance provided by the text prompt. We propose FBSDiff, a novel framework adapting off-the-shelf T2I diffusion model into the I2I paradigm from a fresh frequency-domain perspective. Through dynamic frequency band substitution of diffusion features, FBSDiff realizes versatile and highly controllable text-driven I2I in a plug-and-play manner (without need for model training, fine-tuning, or online optimization), allowing appearance-guided, layout-guided, and contour-guided I2I translation by progressively substituting low-frequency band, mid-frequency band, and high-frequency band of latent diffusion features, respectively. In addition, FBSDiff flexibly enables continuous control over I2I correlation intensity simply by tuning the bandwidth of the substituted frequency band. To further promote image translation efficiency, flexibility, and functionality, we propose FBSDiff++ which improves upon FBSDiff mainly in three aspects: (1) accelerate inference speed by a large margin (8.9$\times$ speedup in inference) with refined model architecture; (2) improve the Frequency Band Substitution module to allow for input source images of arbitrary resolution and aspect ratio; (3) extend model functionality to enable localized image manipulation and style-specific content creation with only subtle adjustments to the core method. Extensive qualitative and quantitative experiments verify superiority of FBSDiff++ in I2I translation visual quality, efficiency, versatility, and controllability compared to related advanced approaches. 
	}

	
	
	
	\maketitle
	
	\section{Introduction}\label{sec1}
	
	I2I translation (abbreviated as I2I hereafter) refers to a class of computer vision tasks that map an input image from one domain to a corresponding output image in another domain. Pix2Pix \cite{isola2017image} pioneers I2I using a supervised conditional GAN \cite{goodfellow2014generative} model trained with domain-paired image data. Subsequently, a series of methods extend I2I to unsupervised paradigm via cycle-consistency constraint \cite{zhu2017unpaired}, geometry-consistent constraint \cite{fu2019geometry}, shared domain subspace \cite{liu2017unsupervised}, and contrastive learning \cite{park2020contrastive}, etc. By obviating the dependency on paired training data, these methods significantly facilitate I2I and catalyze a proliferation in related applications such as makeup transfer \cite{sun2024content,xiang2022ramgan}, painting stylization \cite{gao2020rpd,gao2025sragan}, image cartoonization \cite{jiang2023scenimefy, gao2022learning}, semantic image synthesis \cite{park2019semantic,zhu2020sean}, etc.
	
	Since the advent of CLIP \cite{radford2021learning} aligning vision and language with large-scale contrastive learning, methods have been explored to instruct I2I with text. For example, VQGAN-CLIP \cite{crowson2022vqgan} and FlexIT \cite{couairon2022flexit} translate a source image as per a text by optimizing the encoded latent embedding of the source image with CLIP (image-text similarity) loss. DiffusionCLIP \cite{kim2022diffusionclip} and DiffuseIT \cite{kwon2023diffusion} use CLIP loss to tune the reverse process of the diffusion model \cite{ho2020denoising} for text-guided I2I. Though fall short in visual quality and inference speed, these methods generalize I2I from limited domains to language-guided open domain based on CLIP prior.
	
	Recent advanced methods universally leverage the immense generative prior of the pretrained T2I diffusion models for high-quality text-driven I2I. Initialized from the Latent Diffusion Model (LDM) \cite{rombach2022high}, InstructPix2Pix \cite{brooks2023instructpix2pix} trains a text-guided I2I mapping based on elaborately curated large-scale paired training data for text-instructed image manipulation. IC-Edit \cite{zhang2025context} improves model performance at lower cost by training LoRA adapters of the large T2I model. SINE \cite{zhang2023sine} and Imagic \cite{kawar2023imagic} correlate T2I generation result with an input source image by finetuning the pretrained T2I diffusion model to fit the source image. These methods are suboptimal in efficiency since they either necessitate cost-prohibitive data collection and model training, or require separate model finetuning for each time of model inference. Therefore, recent attention in text-driven I2I has been increasingly focused on inversion-based methods.
	
	Inversion-based methods invert a source image into the Gaussian noise space of the pretrained T2I diffusion model followed by sampling a translated image from the inverted noise guided by the text. To tightly correlate source and generated images, this type of methods focuses on improving inversion accuracy through fixed-point iteration \cite{pan2023effective,garibi2024renoise} or step-wise optimization \cite{mokady2023null,dong2023prompt,li2023stylediffusion}, or attends to designing intricate attention modulation pipeline to ensure I2I structure consistency \cite{parmar2023zero,tumanyan2023plug,cao2023masactrl}. Besides, efforts have also been devoted to I2I efficiency by accelerating the inversion speed \cite{miyake2025negative,nguyen2025swiftedit,samuel2024lightning} as well as I2I text fidelity by boosting the editability of the inverted noise \cite{li2024source,huberman2024edit,kang2025editable}. Despite achieving leading performance, existing inversion-based methods rely on meticulously crafted cross-attention \cite{mokady2023null,dong2023prompt,li2023stylediffusion,miyake2025negative,parmar2023zero} or self-attention \cite{tumanyan2023plug,cao2023masactrl} control pipeline to guarantee I2I correlation, which makes the sampling algorithm unconcise and sensitive to the specific architecture of the denoising network. Moreover, all these methods need to invert the source image conditioned on a manually designed source text, where the source text is required to precisely describe the source image in a format with word-to-word alignment to the target text prompt. This causes inconvenience, especially when the source image is difficult to be textually described.
	
	In this paper, we proposes FBSDiff, a novel inversion-based text-driven I2I method based on a fresh frequency-domain perspective: dynamic frequency band substitution of diffusion features. FBSDiff constructs a pair of parallel diffusion trajectories: reconstruction trajectory (reconstruct source image from the inverted noise vector) and sampling trajectory (randomly generate an image guided by a text). By dynamically transplanting a certain frequency band from the latent diffusion features in the reconstruction trajectory into the corresponding features in the sampling trajectory along the sampling process, a specific form of visual correlation between the source image and the generated image is established, and thus realizing text-driven I2I in a plug-and-play manner.
	
	FBSDiff improves upon related text-driven I2I approaches in model conciseness, versatility, diversity, and controllability.
	
	\textbf{Conciseness}: FBSDiff correlates input and output images by sharing frequency band of latent diffusion features, bypassing the procedurally intricate attention modulation pipeline. This means that our method dispenses with the need to access and manipulate intermediate features inside the denoising network, providing a model-agnostic concise approach for text-driven I2I. Such attention-free property also eliminates the need for the manually designed source text, thus remarkably facilitates I2I especially when the source image is difficult to be textually expressed.

	\textbf{Versatility}: FBSDiff possesses inherent strengths in decomposing and controlling diverse I2I correlations. By virtue of the Frequency Band Substitution (FBS) module, FBSDiff realizes appearance-guided, layout-guided, and contour-guided I2I simply by substituting low-frequency band, mid-frequency band, and high-frequency band of latent diffusion features in the 2D Discrete Cosine Transform (DCT) spectral space, suited to versatile I2I applications. 
	
	\textbf{Diversity}: FBSDiff allows sampling diverse I2I results conditioned on the same source image and text prompt, as opposed to existing advanced methods that are limited to unique I2I result.
	
	\textbf{Controllability}: besides controllability over I2I correlations with different modes of frequency band substitution, FBSDiff also enables continuous control over I2I correlation intensity by tuning the bandwidth of the substituted frequency band.
	
	To further improve model efficiency, flexibility, and functionality, we have developed an advanced version of FBSDiff, termed FBSDiff++. 
	
	\textbf{In efficiency}: FBSDiff++ dispenses with the reconstruction trajectory by extracting guidance features from the inversion trajectory, thus the sampling speed is noticeably accelerated. By avoiding source image reconstruction, FBSDiff++ enables high-quality I2I without pursuing inversion accuracy, which allows for drastically reduced inversion steps than FBSDiff. Therefore, FBSDiff++ improves upon FBSDiff in sampling speed and inversion speed by a large margin.
	
	\textbf{In flexibility}: In contrast to FBSDiff that substitutes feature frequency band in 2D-DCT spectral space, FBSDiff++ transplants frequency band in cascaded 1D-DCT spaces. This refinement eliminates the square-image limitation of FBSDiff, supporting input images of arbitrary aspect ratio. In addition, FBSDiff++ replaces the absolute DCT filtering thresholds adopted by FBSDiff with the percentile-based relative thresholds that lie in a normalized range of [0, 100], further improving user friendliness and flexibility by making the model adaptive to input source images of arbitrary resolution. 
	
	\textbf{In functionality}: FBSDiff++ enables localized image manipulation and style-specific content creation for input images of arbitrary size with only subtle modifications to the core method.
	
	Our main contributions are summarized below:
	\begin{itemize}
		\item We propose FBSDiff, a plug-and-play method adapting pretrained T2I diffusion model to the realm of text-driven I2I from a fresh frequency-domain perspective, realizing high-quality I2I with flexible controllability both in I2I correlation type and correlation intensity.
		\item FBSDiff exhibits noticeable advantages in conciseness and efficiency by obviating model training, finetuning, online optimization, source-text dependency, and the procedurally intricate attention modulation pipeline, all while achieving superiority in visual quality, versatility, and controllability than related methods.
		\item We develop FBSDiff++, an advanced version of FBSDiff with streamlined model architecture, substantially faster inference speed, improved usability and flexibility to adapt to input images of arbitrary size and aspect ratio, and extended functionality to support localized image manipulation and style-specific content creation.
	\end{itemize}
	
	This paper expands upon our conference paper \cite{gao2024fbsdiff} both in methodology and experiments. 
	
	\textbf{In methodology}: (\romannumeral1) FBSDiff++ significantly accelerates inference speed by 8.9 times with a refined model architecture; (\romannumeral2) FBSDiff++ replaces 2D-DCT frequency band substitution (FBS) with consecutive 1D-DCT FBS, allowing for input images of arbitrary aspect ratio; (\romannumeral3) FBSDiff++ converts the absolute DCT filtering thresholds to the percentile-based relative ones, enabling the model to be adaptive to input images of arbitrary resolution. (\romannumeral4) FBSDiff++ extends model functionality to enable localized image manipulation and style-specific content creation for input images of arbitrary size.
	
	\textbf{In experiment}: we supplement abundant results of FBSDiff++ including its new applications; demonstrate superiority of FBSDiff++ over FBSDiff in many aspects; substantially expand baseline methods by incorporating more recent advanced text-driven I2I models for comparison.
	
	\begin{figure*}[t]
		\centering
		\includegraphics[width=0.99\textwidth]{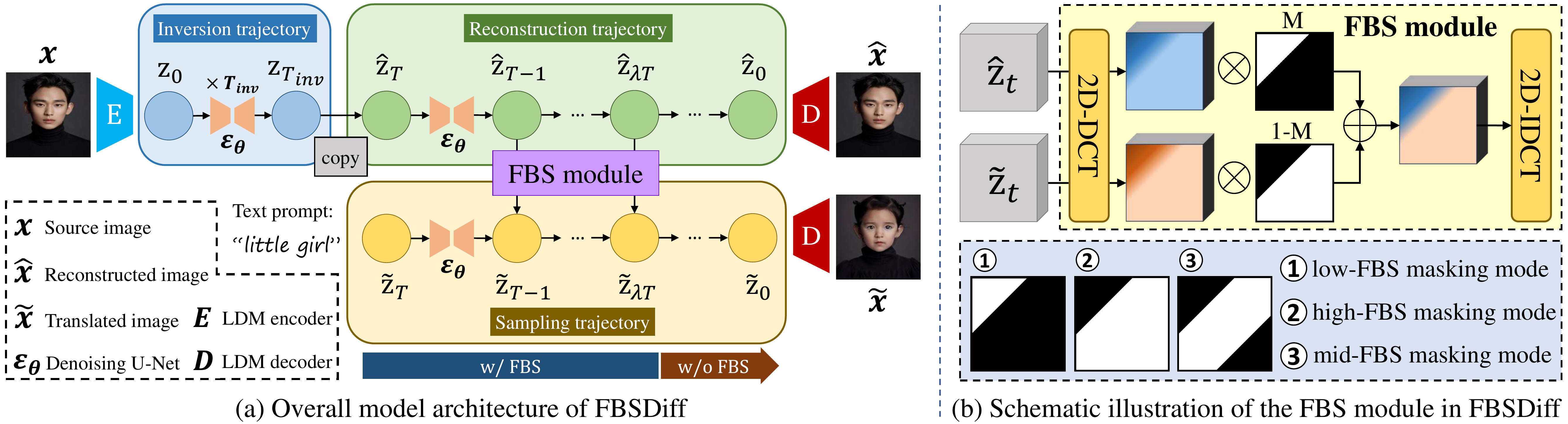}
		\caption{Method overview of FBSDiff (left) as well as illustration of its kernel ingredient: the FBS module (right).}
		\label{fig:FBSDiff_method}
	\end{figure*}
	
	\section{Related work}
	\subsection{Inversion-based text-driven image manipulation}
	Current advanced methods invert source images into the noise space of the pretrained T2I diffusion model for text-driven image manipulation. However, the inversion may not be fully restored due to the local linear assumption of DDIM inversion \cite{song2020denoising}. Accordingly, methods like Renoise \cite{garibi2024renoise}, AIDI \cite{pan2023effective}, and GNRI \cite{samuel2024lightning} boost text-driven image manipulation by leveraging numerical optimization methods to improve inversion accuracy. Another challenge is that the widely used classifer-free guidance (CFG) technique \cite{ho2022classifier} severely breaks the local linear assumption, making the inversion process not reconstructable. This issue is tackled by a series of methods through online optimization. For example, NTI \cite{mokady2023null}, PTI \cite{dong2023prompt}, and StyleDiffusion \cite{li2023stylediffusion} precisely reconstruct the source image under CFG from the inverted noise through per-step optimization over null-text embedding, prompt embedding, and cross-attention \textit{value} embedding, respectively, paving the way for utilizing the cross-attention control pipeline proposed in P2P \cite{hertzprompt} to ensure I2I structure similarity. NPI \cite{miyake2025negative} subtly proposes an optimization-free method to accurately reverse the inversion process under CFG by replacing the null-text embedding in CFG with the prompt embedding. Pix2pix-zero \cite{parmar2023zero} realizes I2I structure consistency by proposing a derivative-based cross-attention control algorithm. Besides focusing on cross-attention control, PnP \cite{tumanyan2023plug} preserves source image structure through dynamic injection of features and self-attention maps inside the denoising network. Masactrl \cite{cao2023masactrl} designs a mutual self-attention framework that generates visually correlated I2I results by aggregating visual information from source images via self-attention layers. As opposed to all these approaches, our method bypasses the need for the prepared source text and the intricate attention manipulation pipeline, all while achieving versatile and highly controllable text-driven I2I.
	
	\subsection{Computer vision in frequency perspective}
	Deep neural network models are mostly applied to tackle vision tasks in spatial or temporal domain, whereas research reveals that model performance can also be boosted from frequency domain. For example, Ghosh et al. \cite{ghosh2016deep} introduce DCT to CNN to accelerate network convergence speed. Xie et al. \cite{xie2021learning} propose a frequency-aware dynamic network for lightweight image super-resolution. Cai et al. \cite{cai2021frequency} impose Fourier frequency spectrum consistency constraint to image translation task for improved identity preservation. FreeU \cite{si2024freeu} improves T2I generation quality by selectively enhancing or depressing different spectral components of diffusion features. Similar idea is also inherited by FreeStyle \cite{he2024freestyle} for text-driven image stylization. This paper combines frequency-domain signal processing with large-scale T2I diffusion model for plug-and-play text-driven I2I.
	
	\section{Method}
	This section introduces FBSDiff and our extended method FBSDiff++. We place related preliminary background in the \textbf{supplementary materials}.
	
	\subsection{FBSDiff}
	\noindent\textbf{Overall model architecture} \newline
	\noindent FBSDiff extends the pretrained LDM from T2I generation to versatile and highly controllable text-driven I2I with a plug-and-play frequency band substitution mechanism. As shown in Fig. \ref{fig:FBSDiff_method}-(a), FBSDiff comprises three diffusion trajectories: (\romannumeral1) inversion trajectory ($z_{0}\rightarrow z_{T_{inv}}$); (\romannumeral2) reconstruction trajectory ($\hat{z}_{T}\rightarrow \hat{z}_{0}$); (\romannumeral3) sampling trajectory (${\tilde{z}}_{T}\rightarrow {\tilde{z}}_{0}$). Starting from the initial feature $z_{0}=E(x)$ extracted from the source image $x$ by the LDM encoder $E$, a $T_{inv}$-step inversion trajectory is built to project $z_{0}$ into the Gaussian noise space conditioned on the null-text embedding $v_{\emptyset}$ via DDIM inversion \cite{song2020denoising}: 
	\begin{equation}
		z_{t+1}=\sqrt{\bar{\alpha}_{t+1}}f_{\theta}(z_{t}, t, v_{\emptyset})+\sqrt{1-\bar{\alpha}_{t+1}}\epsilon_{\theta}(z_{t}, t, v_{\emptyset}),
		\label{eq:inversion}
	\end{equation}
	\begin{equation}
		f_{\theta}(z_{t}, t, v_{\emptyset})=\frac{z_{t}-\sqrt{1-\bar{\alpha}_{t}}\epsilon_{\theta}(z_{t}, t, v_{\emptyset})}{\sqrt{\bar{\alpha}_{t}}},
		\label{eq:back_to_z0_inversion}
	\end{equation}
	where \{$\bar{\alpha}_{t}$\} are schedule parameters, $\epsilon_{\theta}$ is the denoising network of the pretrained LDM. The inverted Gaussian noise $z_{T_{inv}}$ obtained after $T_{inv}$ inversion steps is initialized as the starting noise of the subsequent reconstruction trajectory, which reconstructs $\hat{z}_{0}\approx z_{0}$ from the inverted noise feature $\hat{z}_{T}=z_{T_{inv}}$ via $T$-step DDIM sampling:
	\begin{equation}
		{\hat{z}}_{t-1}=\sqrt{{\bar{\alpha}}_{t-1}}f_{\theta}({\hat{z}}_{t}, t, v_{\emptyset}) + \sqrt{1-{\bar{\alpha}}_{t-1}}\epsilon_{\theta}({\hat{z}}_{t}, t, v_{\emptyset}),
		\label{eq:recon}
	\end{equation}
	\begin{equation}
		f_{\theta}(\hat{z}_{t}, t, v_{\emptyset})=\frac{\hat{z}_{t}-\sqrt{1-\bar{\alpha}_{t}}\epsilon_{\theta}(\hat{z}_{t}, t, v_{\emptyset})}{\sqrt{\bar{\alpha}_{t}}}.
		\label{eq:back_to_z0_recon}
	\end{equation}
	The length of the reconstruction trajectory could be much smaller than that of the inversion trajectory (i.e., $T \ll T_{inv}$) to save inference time. The reconstruction trajectory is conditioned on the same null-text embedding $v_{\emptyset}$ as the inversion trajectory to ensure feature reconstructability (i.e., $\hat{z}_{0}\approx z_{0}$). Meanwhile, an equal-length sampling trajectory is built in parallel with the reconstruction trajectory for T2I generation. The sampling trajectory is also a $T$-step DDIM sampling process that gradually denoises a random Gaussian noise ${\tilde{z}}_{T} \sim \mathcal{N}(0, I)$ into a clean feature ${\tilde{z}}_{0}$ conditioned on the textual embedding $v$ of the target text prompt. To amplify the text guidance effect, we employ the CFG technique \cite{ho2022classifier} by interpolating conditional (target text) and unconditional (null text) noise prediction at each time step with a guidance scale $\omega$ along the sampling trajectory:
	\begin{equation}
		{\tilde{z}}_{t-1}=\sqrt{{\bar{\alpha}}_{t-1}}f^{*}_{\theta}({\tilde{z}}_{t}, t, v) + \sqrt{1-{\bar{\alpha}}_{t-1}}\epsilon^{*}_{\theta}({\tilde{z}}_{t}, t, v),
		\label{eq:sampling}
	\end{equation}
	\begin{equation}
		f^{*}_{\theta}({\tilde{z}}_{t}, t, v)=\frac{{\tilde{z}}_{t}-\sqrt{1-{\bar{\alpha}}_{t}}\epsilon^{*}_{\theta}({\tilde{z}}_{t}, t, v)}{\sqrt{{\bar{\alpha}}_{t}}},
	\end{equation}
	\begin{equation}
		\epsilon^{*}_{\theta}({\tilde{z}}_{t}, t, v)=\omega\cdot \epsilon_{\theta}({\tilde{z}}_{t}, t, v)+(1-\omega)\cdot\epsilon_{\theta}({\tilde{z}}_{t}, t, v_{\emptyset}).
	\end{equation}
	The denoised feature $\tilde{z}_{0}$ is decoded to the output image $\tilde{x}$ via the LDM decoder $D$, i.e., $\tilde{x}=D(\tilde{z}_{0})$. 
	
	To visually correlate the source image $x$ and the output image $\tilde{x}$, we propose a training-free Frequency Band Substitution (FBS) module. By simply inserting the FBS module in between the reconstruction trajectory and the sampling trajectory, versatile and highly controllable text-driven I2I is allowed in a plug-and-play manner. As Fig. \ref{fig:FBSDiff_method}-(a) shows, FBS module dynamically substitutes a certain frequency band of $\tilde{z}_{t}$ with the corresponding frequency band of $\hat{z}_{t}$ along the sampling process. Separated by time step $\lambda T$, the sampling trajectory is divided into two sections. In the early sectioin ($\tilde{z}_{T}\rightarrow \tilde{z}_{\lambda T}$), FBS is applied at each time step to smoothly calibrate the sampling trajectory toward establishing a certain I2I correlation. In the later section ($\tilde{z}_{\lambda T} \rightarrow \tilde{z}_{0}$), FBS is removed to ensure high visual quality of the generated image.
	
	\noindent \textbf{FBS module} \newline
	As illustrated in Fig. \ref{fig:FBSDiff_method}-(b), FBS module takes in a pair of diffusion features $\hat{z}_{t}$ and $\tilde{z}_{t}$, substitutes a certain frequency band of $\tilde{z}_{t}$ with the same frequency band of $\hat{z}_{t}$, and outputs the updated $\tilde{z}_{t}$. 
	
	Firstly, 2D-DCT is utilized to convert input features into DCT space, obtaining their DCT spectra $\hat{S}_{t}=DCT_{2D}(\hat{z}_{t})$, $\tilde{S}_{t}=DCT_{2D}(\tilde{z}_{t})$. The 2D-DCT transformation is formulated as:
	\begin{equation}
		\begin{aligned}
			S^{(n)}_{u,v}=&\frac{2}{\sqrt{hw}}m(u)m(v)\sum\nolimits_{i=0}^{h-1}\sum\nolimits_{j=0}^{w-1}[z^{(n)}_{i,j} \\
			&\cos(\frac{(2i+1)u\pi}{2h})\cos(\frac{(2j+1)v\pi}{2w})],
			\label{eq:2D-DCT}
		\end{aligned}
	\end{equation}
	where $z$ denotes an arbitrary input feature; $S$ denotes the transformed 2D-DCT spectrum; $i,j$ and $u,v$ are 2D coordinate indices of the spatial domain and the 2D-DCT spectral domain, respectively; $h$ and $w$ are height and width of the input feature $z$; $z^{(n)}$ and $S^{(n)}$ refer to the $n$-th channel of $z$ and $S$; $m(0)=\frac{1}{\sqrt{2}}$, $m(\gamma)=1$ for all $\gamma>0$. Then, a certain DCT frequency band of $\hat{S}_{t}$ is extracted and transplanted into $\tilde{S}_{t}$ via a masking operation:
	\begin{equation}
		\tilde{S}^{*}_{t}=\hat{S}_{t}\times M + \tilde{S}_{t} \times (1-M),
		\label{eq:masking}
	\end{equation}
	where $\tilde{S}^{*}_{t}$ is the spectrum of $\tilde{z}_{t}$ after FBS, $M$ is a binary mask designed for extracting and substituting a certain frequency band. Lastly, $\tilde{z}_{t}$ is updated as the result of using 2D-IDCT to convert $\tilde{S}^{*}_{t}$ back to the spatial domain, i.e., $\tilde{z}_{t}=IDCT_{2D}(\tilde{S}^{*}_{t})$. The 2D-IDCT is formulated as:
	\begin{equation}
		\begin{aligned}
			z^{(n)}_{i,j}=&\frac{2}{\sqrt{hw}}\sum\nolimits_{u=0}^{h-1}\sum\nolimits_{v=0}^{w-1}[m(u)m(v)S^{(n)}_{u,v} \\
			&\cos(\frac{(2i+1)u\pi}{2h})\cos(\frac{(2j+1)v\pi}{2w})],
			\label{eq:2D-IDCT}
		\end{aligned}
	\end{equation}
	in which the notation is the same as in Eq. \ref{eq:2D-DCT}. Combining the above-mentioned operations together, the FBS module could be overall formulated as:
	\begin{equation}
		\begin{aligned}
			\tilde{z}_{t} = IDCT_{2D}[&DCT_{2D}(\hat{z}_{t})\times M + \\
			&DCT_{2D}(\tilde{z}_{t}) \times (1-M)].
			\label{eq:FBS_module}
		\end{aligned}
	\end{equation}
	In 2D-DCT spectrum, elements with smaller coordinates (nearer to the top-left origin) encode lower-frequency information while elements with larger coordinates (nearer to the bottom-right corner) correspond to higher-frequency components. Most of the DCT spectral energy is occupied by a small proportion of low-frequency elements near the top-left origin. Different DCT frequency bands correspond to different dimensions of image visual information (e.g., style, structure, layout, contour) and thus could be associated to different I2I correlations in the task of text-driven I2I. 
	
	We accordingly design three modes of frequency band substitution: (\romannumeral1) low-frequency band substitution (low-FBS); (\romannumeral2) high-frequency band substitution (high-FBS); (\romannumeral3) mid-frequency band substitution (mid-FBS). They are respectively realized with three types of binary mask $M$ as used in Eq. \ref{eq:masking}: low-pass filtering mask $M_{lp}$, high-pass filtering mask $M_{hp}$, and mid-pass filtering mask $M_{mp}$, which are specifically defined as follows:
	$$\left\{ 
	\begin{array}{lr}
		M_{lp}(i,j)=1\ if\ i+j \leq th_{lp}\ else\ 0, &  \\
		M_{hp}(i,j)=1\ if\ i+j>th_{hp}\ else\ 0, &  \\
		M_{mp}(i,j)=1\ if\ th_{mp1}<i+j \leq th_{mp2}\ else\ 0,
	\end{array}
	\right.
	$$
	where $i,j$ are 2D coordinate indices of the binary mask. We use the sum of 2D coordinates as the filtering thresholds: $th_{lp}$ is the low-pass filtering threshold; $th_{hp}$ is the high-pass filtering threshold; $th_{mp1}$ and $th_{mp2}$ are lower bound and upper bound of the mid-pass filtering, respectively.
	
	Different modes of Frequency Band Substitution controls different I2I correlations and thus suit versatile text-driven I2I applications:
	\begin{itemize}
		\item For the \textbf{low-FBS} mode switched by using $M_{lp}$, the translated image $\tilde{x}$ is bounded with the source image $x$ in low-frequency visual information which is reflected in image appearance (including the overall style and structure of an image), realizing appearance-guided I2I. This mode especially suits I2I applications pursuing image appearance consistency, e.g., derivative image generation, i.e., derive visually and artistically similar images from a source image.
		\item For the \textbf{high-FBS} mode switched by using $M_{hp}$, the translated image $\tilde{x}$ correlates with the source image $x$ in high-frequency information reflected in image contours, realizing contour-guided I2I. By filtering out low-frequency image style information, this mode especially suits I2I applications pursuing image appearance alteration, e.g., text-driven image style translation.
		\item For the \textbf{mid-FBS} mode switched by using $M_{mp}$, both the low-frequency appearance information and the high-frequency contour information are filtered out, the translated image $\tilde{x}$ correlates with the source image $x$ in mid-frequency layout information, realizing layout-guided I2I.
	\end{itemize}
	
	\begin{algorithm}[t]
		\caption{Complete algorithm of FBSDiff}
		\label{FBSDiff_algorithm}
		\begin{algorithmic}[1]
			\renewcommand{\algorithmicrequire}{\textbf{Input:}}
			\renewcommand{\algorithmicensure}{\textbf{Output:}}
			\REQUIRE{source image $x$, target text prompt.}
			\ENSURE{translated image $\tilde{x}$.}
			\STATE Extract the initial latent feature $z_{0}=E(x)$.
			\FOR{$t=0$ to $T_{inv}-1$}
			\STATE compute $z_{t+1}$ from $z_{t}$ via Eq. \ref{eq:inversion};
			\ENDFOR 
			\COMMENT{DDIM inversion}
			\STATE Initialize $\hat{z}_{T}=z_{T_{inv}}$, $\tilde{z}_{T} \sim \mathcal{N}(0, I)$.
			\FOR{$t=T$ to $\lambda T+1$}
			\STATE compute $\hat{z}_{t-1}$ from $\hat{z}_{t}$ via Eq. \ref{eq:recon};
			\STATE compute $\tilde{z}_{t-1}$ from $\tilde{z}_{t}$ via Eq. \ref{eq:sampling};
			\STATE substitute a frequency band of $\tilde{z}_{t-1}$ with the same frequency band of $\hat{z}_{t-1}$ via Eq. \ref{eq:FBS_module};
			\ENDFOR\COMMENT{DDIM sampling w/ FBS}
			\FOR{$t=\lambda T$ to $1$}
			\STATE compute $\tilde{z}_{t-1}$ from $\tilde{z}_{t}$ via Eq. \ref{eq:sampling};
			\ENDFOR\COMMENT{DDIM sampling w/o FBS}
			\STATE Obtain the final translated image $\tilde{x}=D(\tilde{z}_{0})$.
		\end{algorithmic}
	\end{algorithm}
	
	\begin{figure*}[t]
		\centering
		\includegraphics[width=0.99\textwidth]{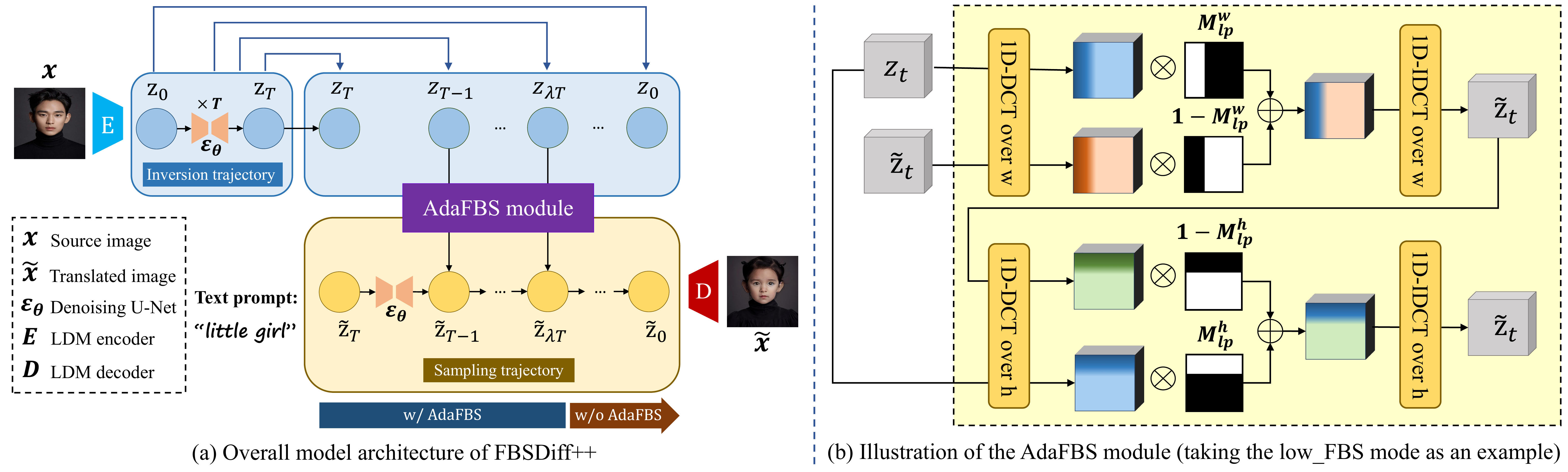}
		\caption{Method overview of FBSDiff++ (left) as well as illustration of its kernel ingredient: the AdaFBS module (right).}
		\label{fig:FBSDiff++_method}
	\end{figure*}
	
	\noindent \textbf{Implementation details} \newline
	We use the pretrained SD v1.5 as the backbone diffusion model and set the CFG scale $\omega=7.5$. We apply 1000-step DDIM inversion ($T_{inv}$=1000) to ensure high-quality image reconstruction, and use 50-step DDIM sampling ($T$=50) for both the reconstruction and the sampling trajectory. We apply FBS only in the first half of the sampling process ($\lambda$=0.5). For the design of the DCT filtering mask, we set $th_{lp}$=80 for $M_{lp}$ in the mode of low-FBS; $th_{hp}$=5 for $M_{hp}$ in the mode of high-FBS; $th_{mp1}$=5, $th_{mp2}$=80 for $M_{mp}$ in the mode of mid-FBS. The complete algorithm of FBSDiff is summarized in Alg. \ref{FBSDiff_algorithm}.
	
	\subsection{FBSDiff++}
	FBSDiff++ extends upon FBSDiff in three aspects: (\romannumeral1) improves model architecture to enable dramatically accelerated inference speed; (\romannumeral2) improves model flexibility and usability by upgrading the FBS module to the AdaFBS module; (\romannumeral3) expand model functionality to allow for localized image manipulation and style-specific content creation for input images of arbitrary size. 
	
	\noindent\textbf{Streamlined model architecture} \newline
	As shown in Fig. \ref{fig:FBSDiff++_method}-(a), FBSDiff++ builds a $T$-step inversion trajectory, inverting the encoded source image embedding $z_{0}=E(x)$ to the Gaussian noise $z_{T}$ conditioned on the null-text embedding $v_{\emptyset}$ as formulated by Eq. \ref{eq:inversion}. The sampling trajectory is a $T$-step DDIM sampling process guided by the target text embedding $v$, which is the same as FBSDiff as given by Eq. \ref{eq:sampling}. FBSDiff++ reverses the order of latent diffusion features in the inversion trajectory to construct the guiding trajectory ($z_{T}\rightarrow z_{0}$) in parallel with the sampling trajectory ($\tilde{z}_{T}\rightarrow \tilde{z}_{0}$). Such design bypasses the $T$-step DDIM sampling process of the original reconstruction trajectory, leading to noticeably accelerated sampling speed. In addition, by eliminating the need to reconstruct the source image, the demand for the inversion accuracy is dramatically relaxed, which allows for substantially fewer inversion steps ($T$) than FBSDiff ($T_{inv} \geq T$), thus significantly reduces the time cost of the inversion process. Lastly, versatile and highly controllable I2I is realized by plugging in our improved AdaFBS module in between the guiding trajectory and the sampling trajectory, where AdaFBS is applied only in the early section of the sampling trajectory separated by the time step $\lambda T$. The complete algorithm of FBSDiff++ is summarized in Alg. \ref{FBSDiff++_algorithm}.
	
	\begin{algorithm}[t]
		\caption{Complete algorithm of FBSDiff++}
		\label{FBSDiff++_algorithm}
		\begin{algorithmic}[1]
			\renewcommand{\algorithmicrequire}{\textbf{Input:}}
			\renewcommand{\algorithmicensure}{\textbf{Output:}}
			\REQUIRE{source image $x$, target text prompt.}
			\ENSURE{translated image $\tilde{x}$.}
			\STATE Extract the initial latent feature $z_{0}=E(x)$.
			\STATE Initialize an empty list $L$.
			\FOR{$t=0$ to $T-1$}
			\STATE compute $z_{t+1}$ from $z_{t}$ via Eq. \ref{eq:inversion};
			\STATE $L$.append($z_{t+1}$);
			\ENDFOR 
			\COMMENT{DDIM inversion}
			\STATE $L$.pop(-1). \COMMENT{remove the last element $z_{T}$}
			\STATE Initialize $\tilde{z}_{T} \sim \mathcal{N}(0, I)$.
			\FOR{$t=T$ to $\lambda T+1$}
			\STATE compute $\tilde{z}_{t-1}$ from $\tilde{z}_{t}$ via Eq. \ref{eq:sampling};
			\STATE $z_{t-1}$ = $L$.pop(-1);
			\STATE $\tilde{z}_{t-1}$ = AdaFBS($z_{t-1}$,$\tilde{z}_{t-1}$); (Fig. \ref{fig:FBSDiff++_method}-(b))
			\ENDFOR\COMMENT{DDIM sampling w/ AdaFBS}
			\FOR{$t=\lambda T$ to $1$}
			\STATE compute $\tilde{z}_{t-1}$ from $\tilde{z}_{t}$ via Eq. \ref{eq:sampling};
			\ENDFOR\COMMENT{DDIM sampling w/o AdaFBS}
			\STATE Obtain the final translated image $\tilde{x}=D(\tilde{z}_{0})$.
		\end{algorithmic}
	\end{algorithm}
	
	\begin{figure}[t]
		\centering
		\includegraphics[width=\linewidth]{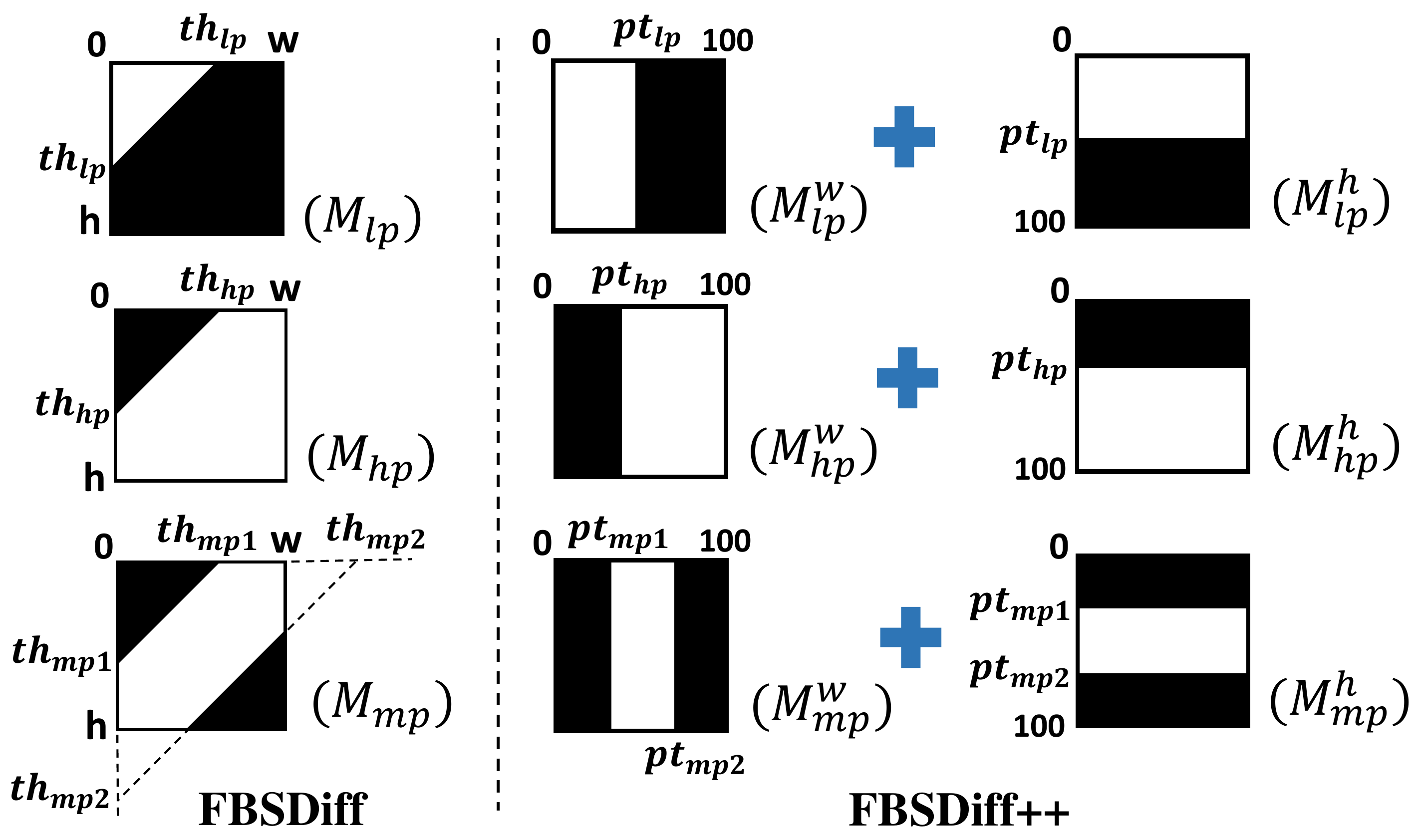}
		\caption{Comparison between the DCT filtering masks used in FBSDiff (left) and FBSDiff++ (right).}
		\label{fig:FBS_vs_AdaFBS}
	\end{figure}
	
	\begin{figure}[t]
		\centering
		\includegraphics[width=\linewidth]{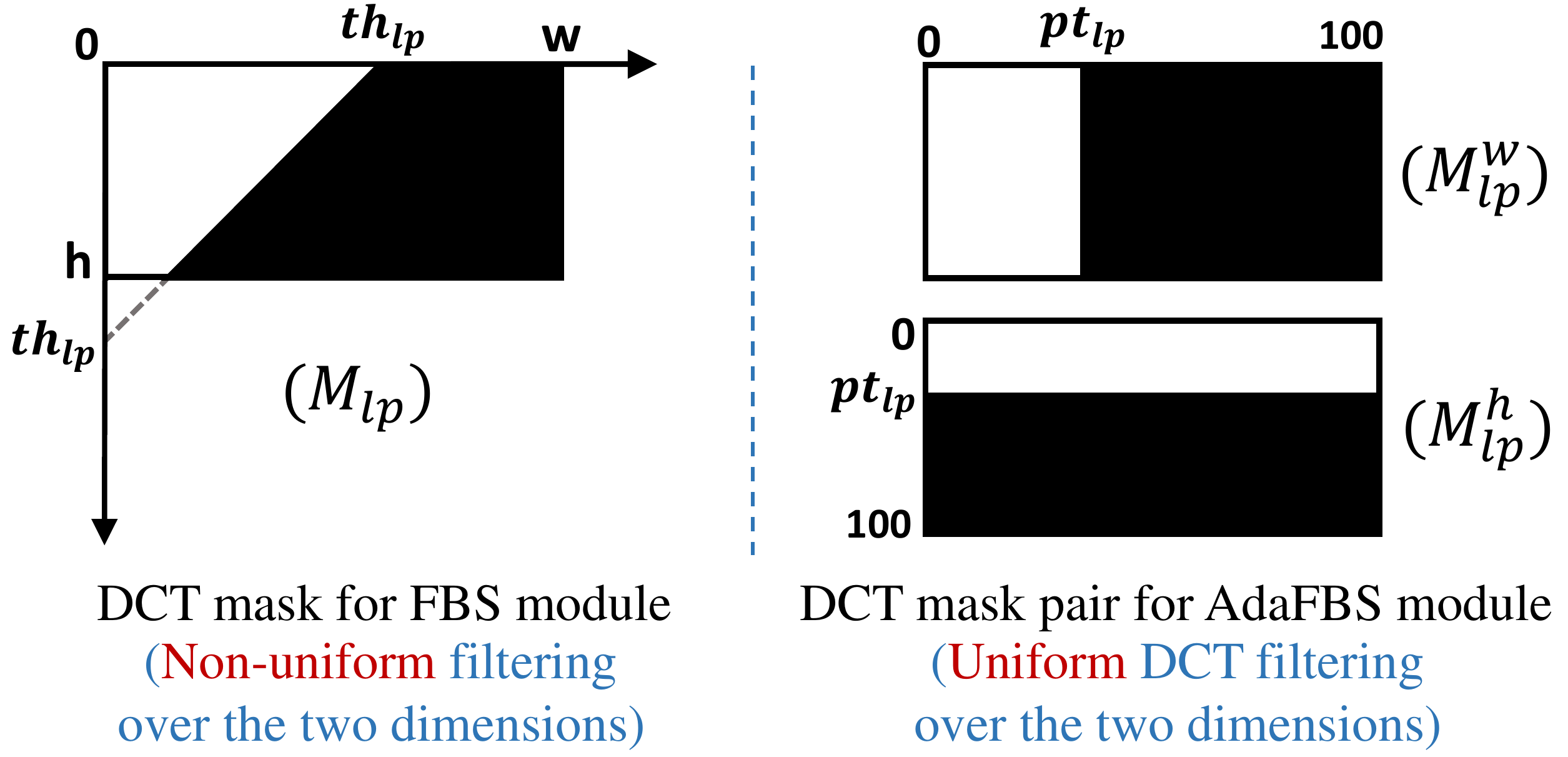}
		\caption{FBS does not allow uniform-strength DCT filtering  over the two spatial dimensions for non-square source images, AdaFBS ensures equal-proportion DCT filtering over the two dimensions for arbitrary image aspect ratio.}
		\label{fig:AdaFBS_merit}
	\end{figure}
	
	\noindent\textbf{AdaFBS module} \newline
	FBSDiff++ adapts to input images with arbitrary aspect ratio and resolution by upgrading FBS module to AdaFBS module. As Fig. \ref{fig:FBS_vs_AdaFBS} illustrates, the key improvements of AdaFBS focus mainly on two aspects: (\romannumeral1) different from FBS that substitutes feature frequency band through 2D DCT masking where the sum of 2D coordinates is used as the masking threshold, AdaFBS adopts two consecutive 1D DCT masking where the filtering threshold of each spatial dimension is adaptively determined to ensure equal-degree frequency band substitution over the two dimensions; (\romannumeral2) AdaFBS employs percentile-based relative values as DCT filtering thresholds rather than the absolute values used by the FBS module. 
	
	The 1D-DCT and 1D-IDCT used in the AdaFBS module are given by Eq. \ref{eq:1D-DCT} and Eq. \ref{eq:1D-IDCT}:
	\begin{equation}
		S^{(n)}_{k}=\sqrt{\frac{2}{L}}\cdot c_{k}\cdot \sum_{l=0}^{L-1}z_{l}^{(n)}\cdot \cos(\frac{\pi k(2l+1)}{2L}),
		\label{eq:1D-DCT}
	\end{equation}
	\begin{equation}
		z^{(n)}_{l}=\sqrt{\frac{2}{L}}\cdot \sum_{k=0}^{L-1}c_{k}\cdot S_{k}^{(n)}\cdot \cos(\frac{\pi k(2l+1)}{2L}),
		\label{eq:1D-IDCT}
	\end{equation}
	where $z$ is the latent diffusion feature, $S$ is the corresponding DCT spectrum, $n$ represents the $n$-th channel, $L$ denotes the sequence length, which is either feature height $h$ or width $w$ depending on which spatial dimension to apply 1D-DCT on.
	
	As illustrated in Fig. \ref{fig:FBS_vs_AdaFBS}, AdaFBS designs three modes of DCT mask pairs ($M_{lp}^{w}$,$M_{lp}^{h}$), ($M_{hp}^{w}$,$M_{hp}^{h}$), ($M_{mp}^{w}$,$M_{mp}^{h}$) to realize low-FBS, high-FBS, and mid-FBS, which are formulated as follows:
	$$\left\{ 
	\begin{array}{lr}
		M_{lp}^{w}(i,j)=1\ if\ j \leq (pt_{lp}\cdot w)/100\ else\ 0, &  \\
		M_{hp}^{w}(i,j)=1\ if\ j>(pt_{hp}\cdot w)/100\ else\ 0, &  \\
		M_{mp}^{w}(i,j)=1\ if\ \frac{pt_{mp1}\cdot w}{100}<j \leq \frac{pt_{mp2}\cdot w}{100}\ else\ 0,
	\end{array}
	\right.
	$$
	$$\left\{ 
	\begin{array}{lr}
		M_{lp}^{h}(i,j)=1\ if\ i \leq (pt_{lp}\cdot h)/100\ else\ 0, &  \\
		M_{hp}^{h}(i,j)=1\ if\ i>(pt_{hp}\cdot h)/100\ else\ 0, &  \\
		M_{mp}^{h}(i,j)=1\ if\ \frac{pt_{mp1}\cdot h}{100}<i \leq \frac{pt_{mp2}\cdot h}{100}\ else\ 0,
	\end{array}
	\right.
	$$
	where $h$ and $w$ denote feature height and width; $i$ and $j$ are coordinate indices in the vertical and horizontal directions; $pt_{lp}$ and $pt_{hp}$ are percentile thresholds of the low-FBS mode and the high-FBS mode; $pt_{mp1}$ and $pt_{mp2}$ are the lower bound and upper bound percentiles for the mid-FBS mode. By switching to percentile-based filtering thresholds, the AdaFBS module can adapt to source images of arbitrary resolution with normalized percentiles ranging in $\left[0, 100\right]$. As Fig. \ref{fig:AdaFBS_merit} displays, the DCT masking of the FBS module could not apply equal-degree filtering over the two spatial dimensions for non-square source images (e.g., the extracted frequency band covers partial spectral range in the $w$ direction but full range in the $h$ direction), while the AdaFBS module achieves uniform-strength DCT filtering over the two spatial dimensions by applying equally proportional 1D-DCT filtering to each spatial dimension separately, and thus flexibly adapts to source images of arbitrary aspect ratio. Fig. \ref{fig:FBSDiff++_method}-(b) takes the low-FBS mode as an example to illustrate the AdaFBS module. The complete program of the AdaFBS module is presented in Alg. \ref{AdaFBS_algorithm}, in which 1D-DCT$_{w}$ and 1D-DCT$_{h}$ denote 1D-DCT transformation over the $w$ and $h$ dimension, respectively, and the same for 1D-IDCT$_{w}$ and 1D-IDCT$_{h}$.
	\begin{algorithm}[t]
		\caption{Algorithm of the AdaFBS module}
		\label{AdaFBS_algorithm}
		\begin{algorithmic}[1]
			\renewcommand{\algorithmicrequire}{\textbf{Input:}}
			\renewcommand{\algorithmicensure}{\textbf{Output:}}
			\REQUIRE{inversion feature $z_{t}$, sampling feature $\tilde{z}_{t}$, DCT mask pair ($M_{*}^{w}$, $M_{*}^{h}$) ($* \in \left[lp, hp, mp\right]$).}
			\ENSURE{updated $\tilde{z}_{t}$ after AdaFBS.}
			\STATE $S_{t}^{w}$=1D-DCT$_{w}$($z_{t}$), $\tilde{S}_{t}^{w}$=1D-DCT$_{w}$($\tilde{z}_{t}$).
			\STATE $\tilde{z}_{t}$=1D-IDCT$_{w}(S_{t}^{w} \times M_{*}^{w} + \tilde{S}_{t}^{w} \times (1-M_{*}^{w}))$ 
			\STATE $S_{t}^{h}$=1D-DCT$_{h}$($z_{t}$), $\tilde{S}_{t}^{h}$=1D-DCT$_{h}$($\tilde{z}_{t}$).
			\STATE $\tilde{z}_{t}$=1D-IDCT$_{h}(S_{t}^{h} \times M_{*}^{h} + \tilde{S}_{t}^{h} \times (1-M_{*}^{h}))$.
		\end{algorithmic}
	\end{algorithm}
	
	\begin{figure*}[t]
		\centering
		\includegraphics[width=0.99\linewidth]{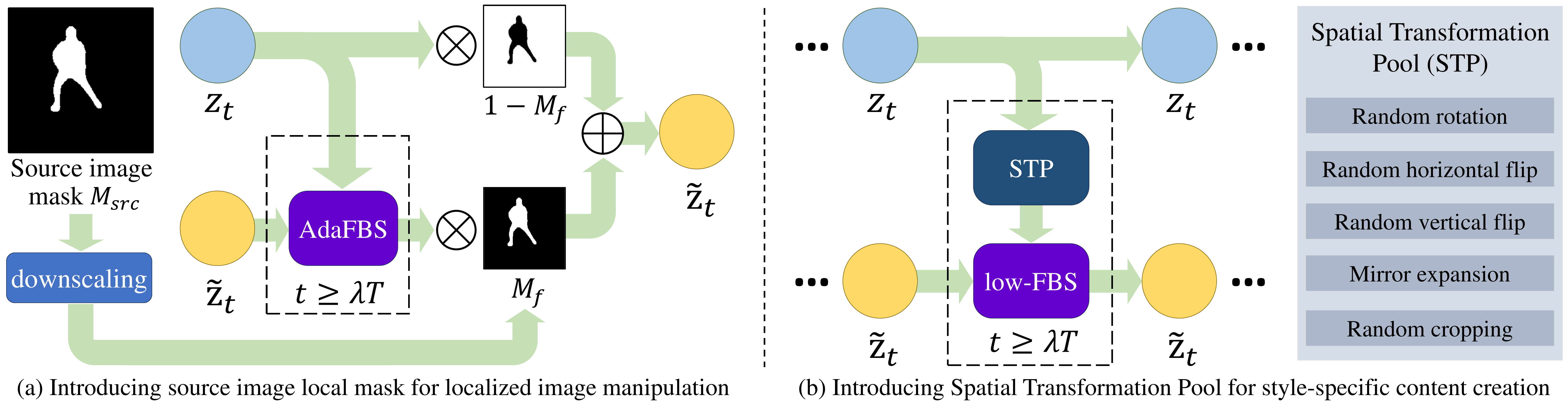}
		\caption{Minor adjustments of FBSDiff++ to cater to localized image manipulation and style-specific content creation.}
		\label{fig:AdaFBS_augment}
	\end{figure*}
	
	\noindent\textbf{Extended functionalities} \label{extended} \newline 
	Similar to FBSDiff, FBSDiff++ allows derivative image generation, image style translation, layout-guided I2I by switching to low-FBS, high-FBS, and mid-FBS, respectively. Besides, FBSDiff++ has been enhanced with new functionalities including localized image manipulation and style-specific content creation simply with minor adjustments to the core method (shown in Fig. \ref{fig:AdaFBS_augment}). 
	
	For localized image manipulation, FBSDiff++ allows to textually manipulate only a local region of the source image as indicated by a binary mask of the source image $M_{src}$. The method adjustment to implement this function is illustrated in Fig. \ref{fig:AdaFBS_augment}-(a). Firstly, the source image mask $M_{src}$ is downscaled to the spatial size of the latent diffusion features, yielding a feature mask $M_{f}$. Then, $M_{f}$ is incorporated into each step of the sampling process for spatial-level feature calibration. The feature update rule during the early section of the sampling trajectory $w/$ AdaFBS is modified as:
	\begin{equation}
		\tilde{z}_{t}=AdaFBS(z_{t},\tilde{z}_{t})\times M_{f}+z_{t}\times (1-M_{f}),
	\end{equation}
	while the update rule during the later section of the trajectory $w/o$ AdaFBS is adjusted to:
	\begin{equation}
		\tilde{z}_{t}=\tilde{z}_{t}\times M_{f}+z_{t}\times (1-M_{f}).
	\end{equation}
	
	For style-specific content creation, FBSDiff++ allows to generate a target image which is faithful to the text prompt in image content while inheriting the style of the source image. The corresponding implementation is illustrated in Fig. \ref{fig:AdaFBS_augment}-(b). To ensure I2I consistency in style pattern, the low-FBS mode is activated in the AdaFBS module. However, low-FBS correlates low-frequency visual information including both image style and global structure, which contradicts the goal of structurally decoupled content creation. To break I2I structural correlation and achieve pure style consistency, we build a Spatial Transformation Pool (STP) and embed it before the AdaFBS (low-FBS) module to spatially perturb the structure of $z_{t}$. The STP is sequentially composed of the following spatial transformation operations. (\romannumeral1) Random rotation: rotate $z_{t}$ with a degree randomly sampled in [$0\degree, 90\degree, 180\degree, 270\degree$]. (\romannumeral2) Random horizontal flip: flip $z_{t}$ horizontally with 50\% probability. (\romannumeral3) Random vertical flip: flip $z_{t}$ vertically with 50\% probability. (\romannumeral4) Mirror expansion: expand $z_{t}$ from the shape of ($h, w$) to ($3h, 3w$) using mirror symmetry to ensure border continuity. (\romannumeral5) Random cropping: randomly crop a patch with random height in [$h, 3h$] and random width in [$w, 3w$] from the expanded $z_{t}$ followed by resizing to ($h, w$). These procedures randomly distort the spatial arrangement of $z_{t}$ while preserving its style pattern. The rotation degree in step-(\romannumeral1), the boolean value of whether flip or not in step-(\romannumeral2) and step-(\romannumeral3), and the size and location of the cropped patch in step-(\romannumeral5) are all randomly sampled once and kept fixed along the sampling trajectory. 
	
	The full algorithms of the above two functionalities are shown in the supplementary materials.
	
	\noindent\textbf{I2I correlation strength adjustment} \newline
	FBSDiff++ enables continuously controllable I2I correlation intensity simply by tuning the percentile thresholds $pt_{lp}$, $pt_{hp}$, and $(pt_{mp1}$, $pt_{mp2})$ in low-FBS, high-FBS, and mid-FBS mode, respectively, where all percentiles are tunable in the range of [0, 100] regardless of source image size.
	
	\noindent\textbf{Implementation details} \newline
	For FBSDiff++, both the inversion steps and the sampling steps are 50, i.e., $T=50$. Similarly, the AdaFBS module is applied only in the first half of the sampling trajectory, i.e., $\lambda=0.5$. By default, the low-pass percentile $pt_{lp}$ is 60, the high-pass percentile $pt_{hp}$ is 5, and the mid-pass percentiles $(pt_{mp1},pt_{mp2})$ are $(7,50)$.
	
	\begin{figure*}[t]
		\centering
		\includegraphics[width=0.98\linewidth]{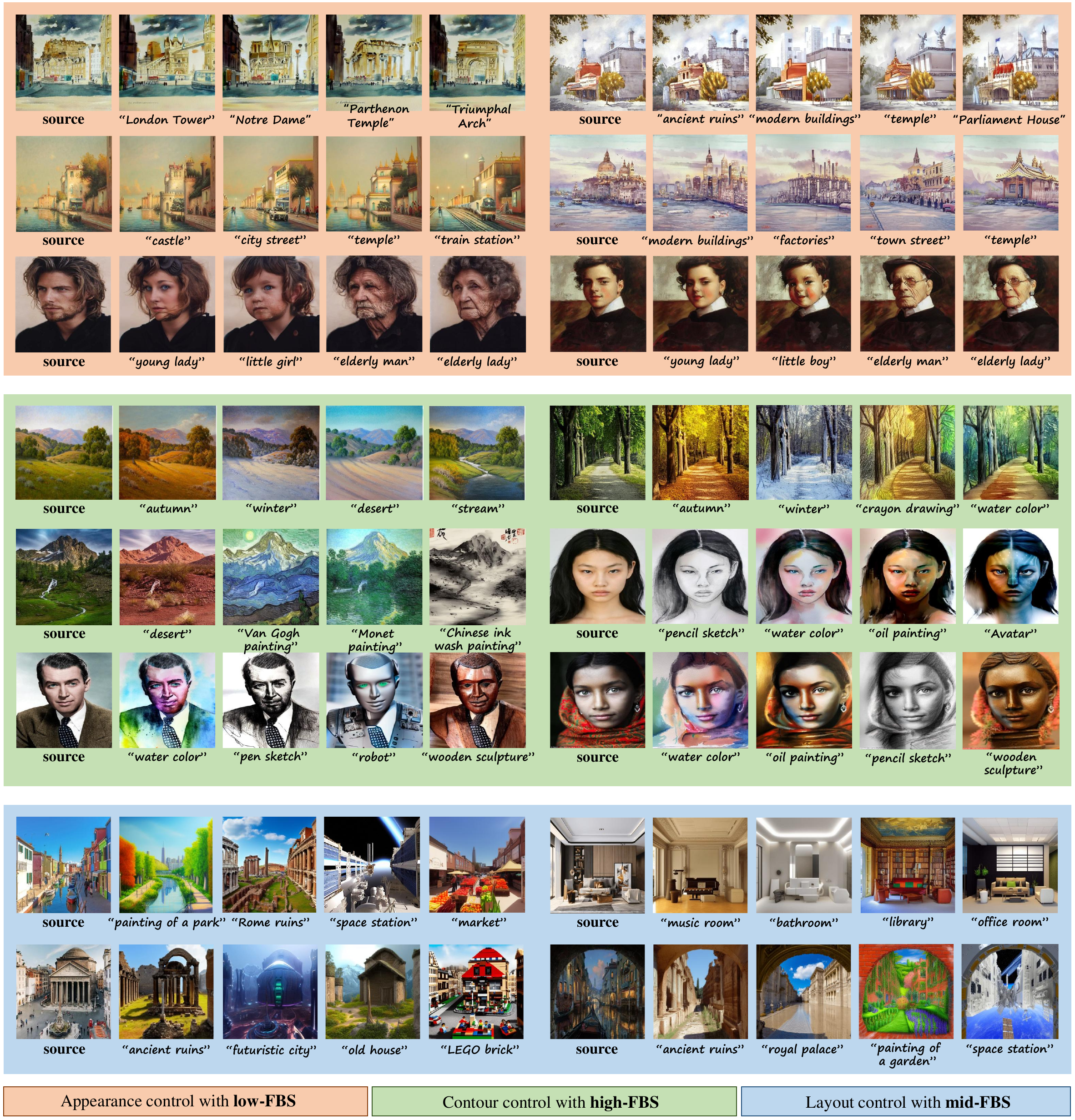}
		\caption{Qualitative results of FBSDiff for different I2I applications, including derivative image generation based on image appearance control in the mode of low-FBS, image style translation based on image contour control in the mode of high-FBS, and layout-guided I2I based on image layout control in the mode of mid-FBS. Results are tested with input source images of $512\times512$ resolution (the default image size of SD v1.5). \textbf{Better viewed with zoom-in}.}
		\label{fig:FBSDiff_qualitative_results}
	\end{figure*}
	
	\begin{figure*}[t]
		\centering
		\includegraphics[width=0.98\linewidth]{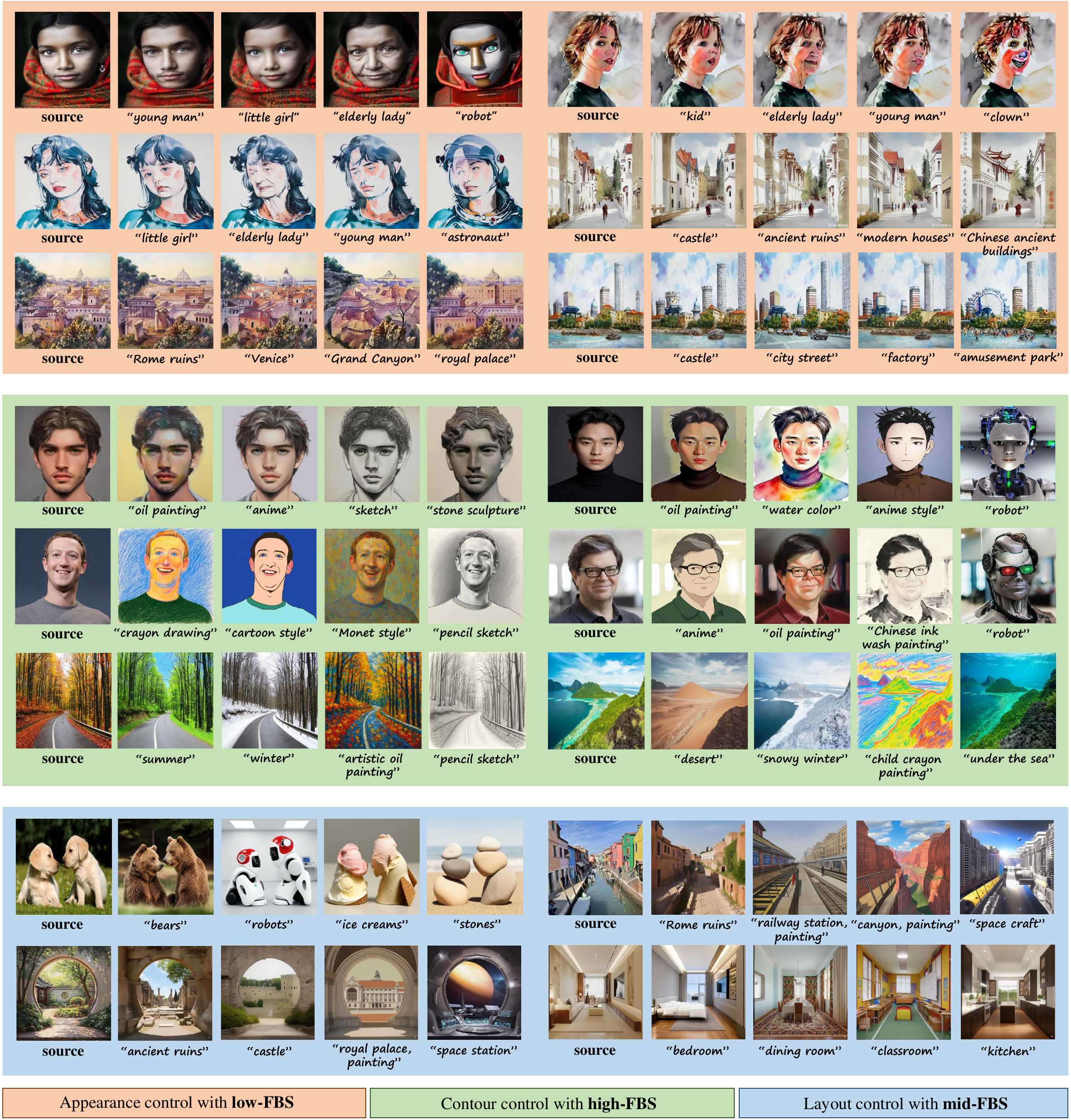}
		\caption{Qualitative results of FBSDiff++ for different I2I applications, including derivative image generation based on image appearance control in the mode of low-FBS, image style translation based on image contour control in the mode of high-FBS, and layout-guided I2I based on image layout control in the mode of mid-FBS. Results are tested with input source images of $512\times512$ resolution (the default image size of SD v1.5). \textbf{Better viewed with zoom-in}.}
		\label{fig:FBSDiff++_qualitative_results}
	\end{figure*}
	
	\begin{figure*}[t]
		\centering
		\includegraphics[width=0.99\linewidth]{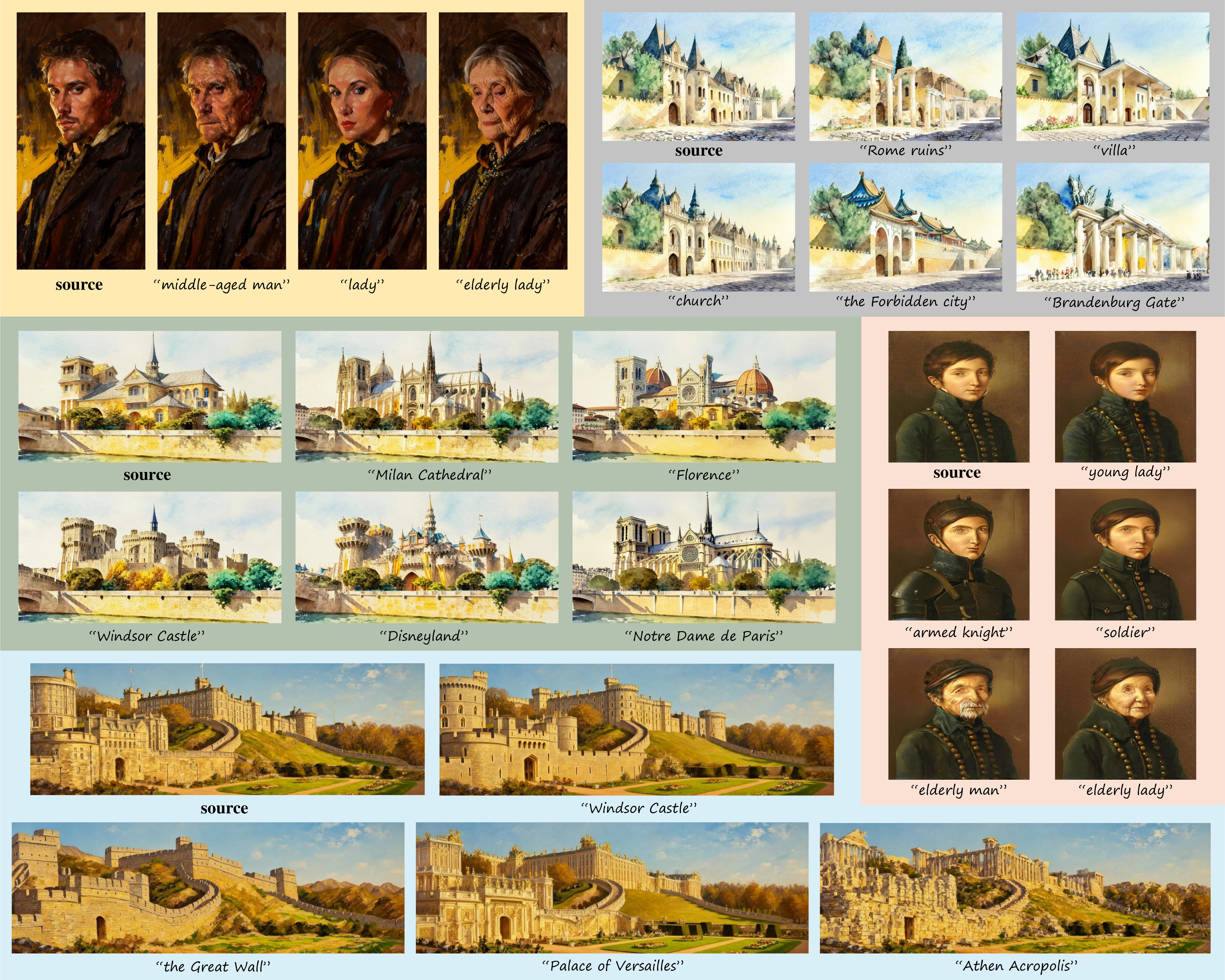}
		\caption{Taking the derivative image generation application in the mode of low-FBS for example, FBSDiff++ enables high-quality I2I for input source images of arbitrary resolution and aspect ratio. Results maintain source image's appearance accurately while comply with the text faithfully. The image resolution in the yellow pannel, gray pannel, green pannel, pink pannel, blue pannel is 1024$\times$512, 512$\times$768, 512$\times$1024, 768$\times$768, 512$\times$1536, respectively. \textbf{Better viewed with zoom-in}.}
		\label{fig:arbitrary_size}
	\end{figure*}

	\section{Experiments}\label{sec2}
	\subsection{Qualitative analyses}
	Fig. \ref{fig:FBSDiff_qualitative_results} displays some example test results of FBSDiff for different text-driven I2I applications. In the mode of low-FBS, by dynamically transplanting feature low-frequency band, the translated images are highly similar to the source image in appearance (including image style and global structure), since image appearance corresponds to low-frequency components of latent diffusion features. This mode suits I2I application to derive visually similar but semantically different images from a same source image template, i.e., derivative image generation. In the mode of high-FBS, the substitution of high-frequency band of diffusion features establishes I2I contour similarity, while the low-frequency style information of the target image is randomly generated according to the text prompt. Such mode suits I2I application to transform image style via a text, i.e., text-driven image style translation. In the mode of mid-FBS, the sharing of mid-frequency band of diffusion features corresponds to I2I layout consistency, applying to the application of layout-guided I2I. Fig. \ref{fig:FBSDiff++_qualitative_results} displays example test results of FBSDiff++. FBSDiff++ also produces visually appealing I2I results with appearance consistency, contour consistency, and layout consistency in the mode of low-FBS, high-FBS, and mid-FBS, respectively, all while achieving high text fidelity.

	\begin{figure*}[t]
		\centering
		\includegraphics[width=0.99\linewidth]{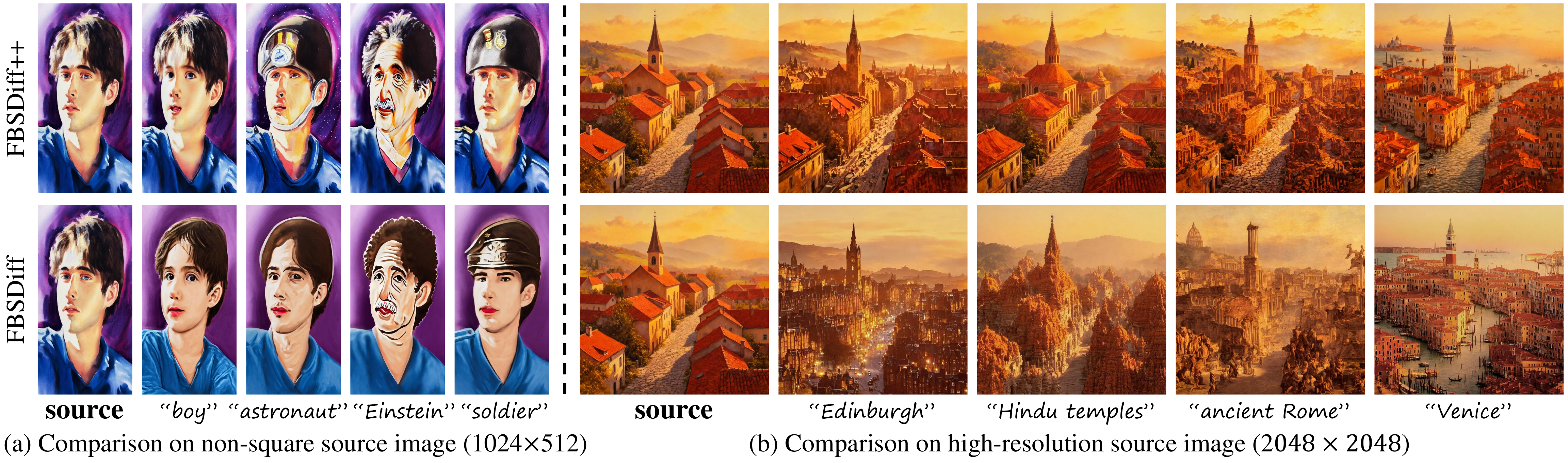}
		\caption{Visual comparison between FBSDiff and FBSDiff++. FBSDiff generates unnatural I2I results for non-square source image due to non-uniform (unequal-proportion) DCT filtering over the two spatial dimensions, and generates visually less correlated I2I results for large-size source image since the DCT filtering of the FBS module is not adaptive to image size. These two issues have been well solved by FBSDiff++ with the improved AdaFBS module. \textbf{Better viewed with zoom-in}.}
		\label{fig:FBSDiff_vs_FBSDiff++}
	\end{figure*}
	
	\begin{figure*}[t]
		\centering
		\includegraphics[width=0.99\linewidth]{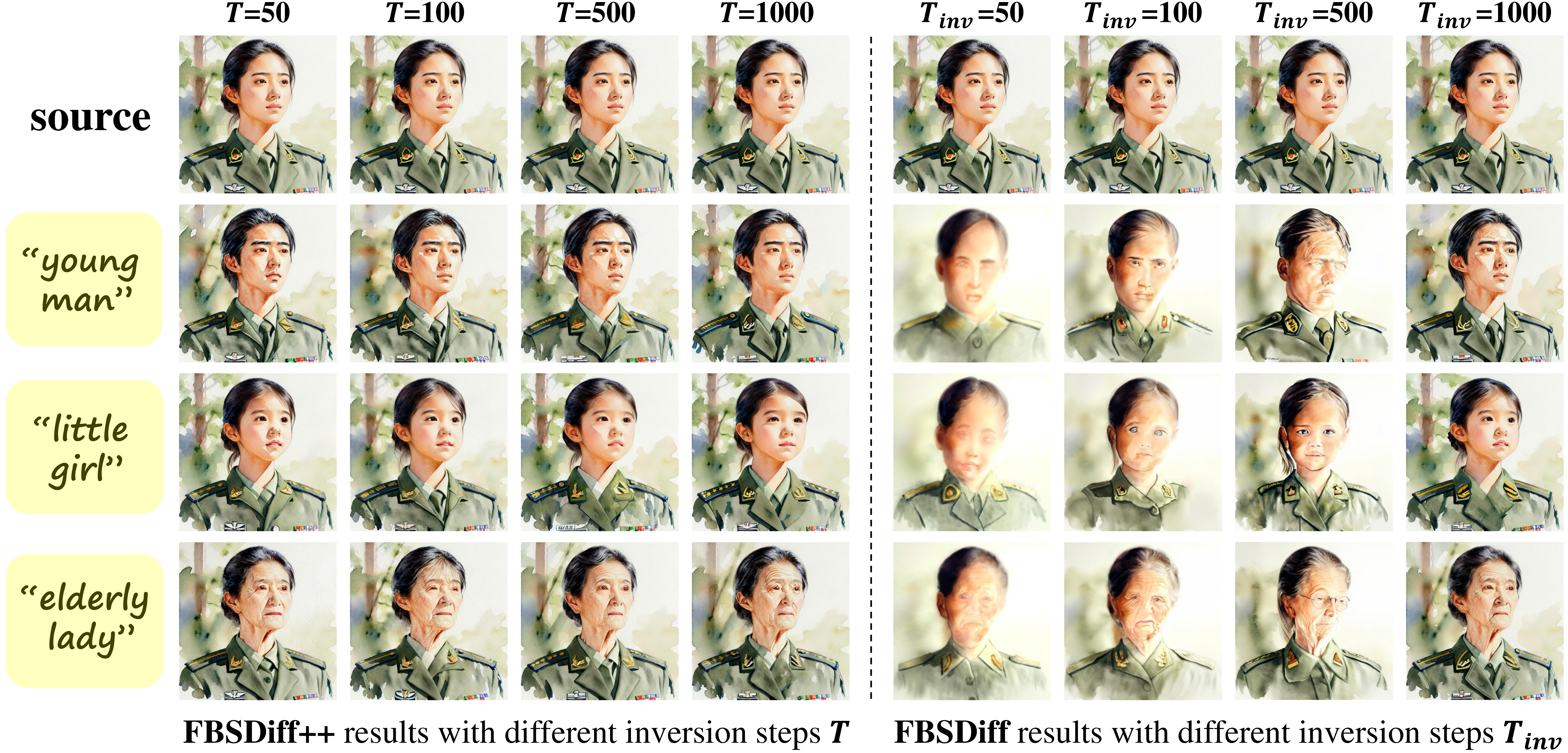}
		\caption{Visual comparison between FBSDiff and FBSDiff++ in different inversion steps. The generation quality of FBSDiff is highly dependent on the reconstruction accuracy of DDIM inversion, which leads to unsatisfactory I2I results with fewer inversion steps (e.g., $T_{inv}=100$). FBSDiff++ bypasses source image reconstruction and enables to produce high-quality results with only 50 inversion steps (e.g., $T=50$), improving inference speed dramatically. \textbf{Better viewed with zoom-in}.}
		\label{fig:FBSDiff_vs_FBSDiff++_in_inversion_steps}
	\end{figure*}
	
	The first significant improvement of FBSDiff++ over FBSDiff is the flexibility of the source image size. As Fig. \ref{fig:arbitrary_size} displays, taking the low-FBS mode for example, FBSDiff++ allows high-quality text-driven I2I for input source image of arbitrary resolution and aspect ratio, rather than the standard $512 \times 512$ resolution (the default image size for SD v1.5) confined by FBSDiff. As visually compared in Fig. \ref{fig:FBSDiff_vs_FBSDiff++}, taking the low-FBS mode as an example again, FBSDiff may generate unnatural I2I results with ruined semantic accuracy for non-square source image (see the left panel). We empirically observe that this is because the 2D-DCT masking designed in the FBS module cannot perform uniform-strength (equal-proportion) DCT filtering over the two spatial dimensions when the source image aspect ratio is not $1:1$. In contrast, FBSDiff++ solves this issue since the AdaFBS module replaces 2D-DCT filtering with two consecutive equally proportional 1D-DCT operations. Moreover, the right panel of Fig. \ref{fig:FBSDiff_vs_FBSDiff++} shows that FBSDiff may generate less correlated I2I results for particularly large source image (e.g., the translated image is less consistent with the source image in style and hue in low-FBS mode). This is due to that the absolute DCT filtering thresholds of FBSDiff are tuned for specific image size ($512 \times 512$), therefore, the substituted spectral components may not be sufficient to maintain high I2I correlation for much larger source image. FBSDiff++ circumvents this problem with percentile-based relative thresholds, which makes the frequency band substitution adaptive to arbitrary image resolution.
	
	\begin{figure}[t]
		\centering
		\includegraphics[width=\linewidth]{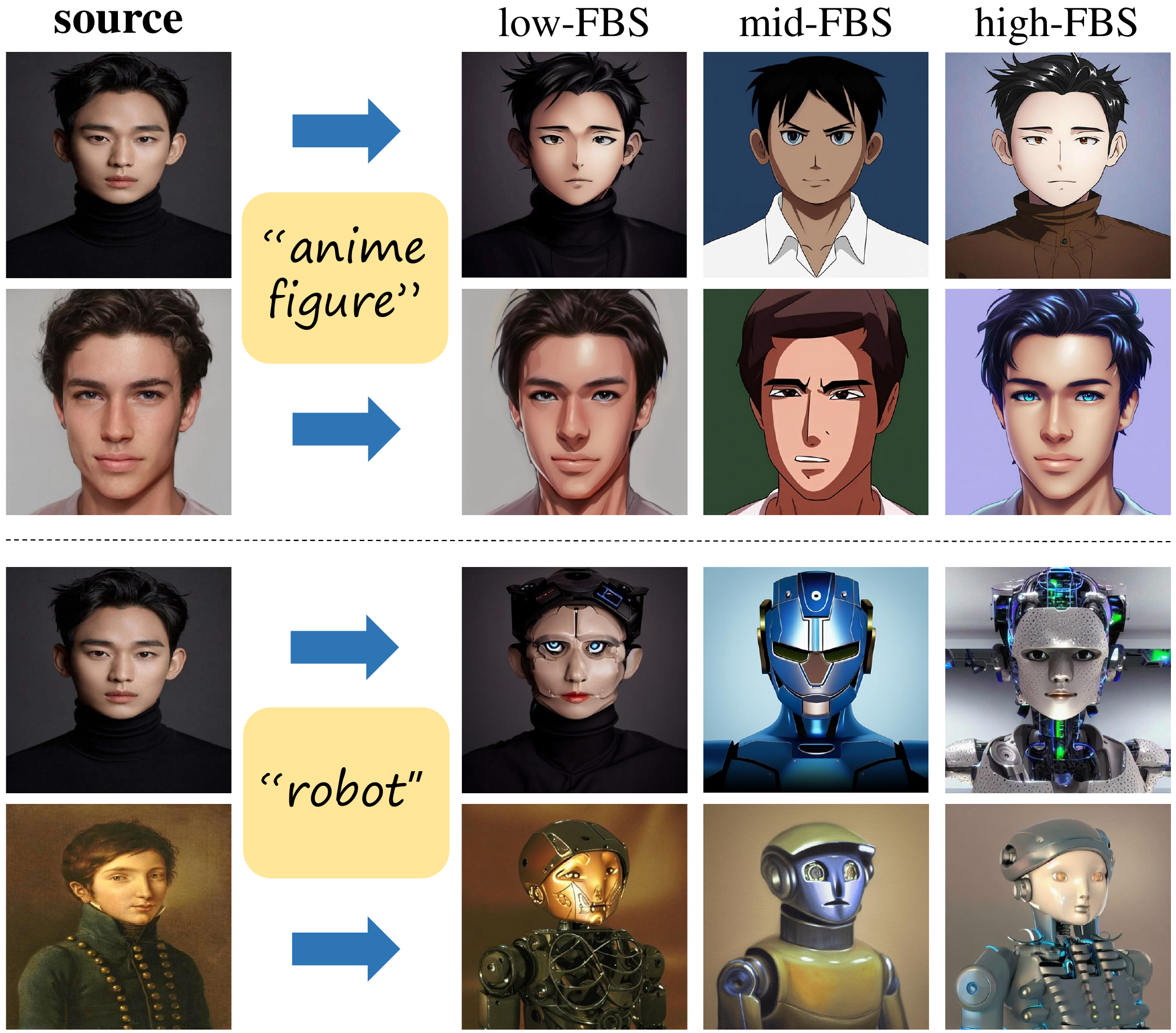}
		\caption{Visual comparison between different modes of AdaFBS. low-FBS maintains image style and global structure, high-FBS preserves image structure while enables significant style change, mid-FBS allows style change and contour deviation while maintains image layout.}
		\label{fig:high_mid_low_compare}
	\end{figure}
	
	The second major enhancement of FBSDiff++ over FBSDiff is inference efficiency. Fig. \ref{fig:FBSDiff_vs_FBSDiff++_in_inversion_steps} presents I2I results of FBSDiff and FBSDiff++ in low-FBS mode with different lengths of the inversion trajectory. FBSDiff fails to produce satisfactory results except for setting a very large number of inversion steps. This is because the generation quality and I2I consistency of FBSDiff heavily depend on the precision of the source image reconstruction, while high reconstruction accuracy requires a sufficiently large number of inversion steps (e.g., $T_{inv}$=1000), leading to overly long inversion trajectory and thus much slower inference speed. In contrast, FBSDiff++ is not sensitive to the length of the inversion trajectory. Results of FBSDiff++ with only 50 inversion steps ($T$=50) are visually on par with those with 1000 inversion steps ($T$=1000). This efficiency advantage is due to the architectural improvement of FBSDiff++. Different from FBSDiff that builds guidance features from the reconstruction trajectory, FBSDiff++ removes the reconstruction trajectory and extracts guidance features along the reverse direction of the inversion trajectory. By dispensing with source image reconstruction, FBSDiff++ does not require sufficiently long inversion trajectory to ensure reconstruction accuracy, and thus remarkably accelerates inference speed with much fewer inversion steps. 
	
	\begin{figure}[t]
		\centering
		\includegraphics[width=\linewidth]{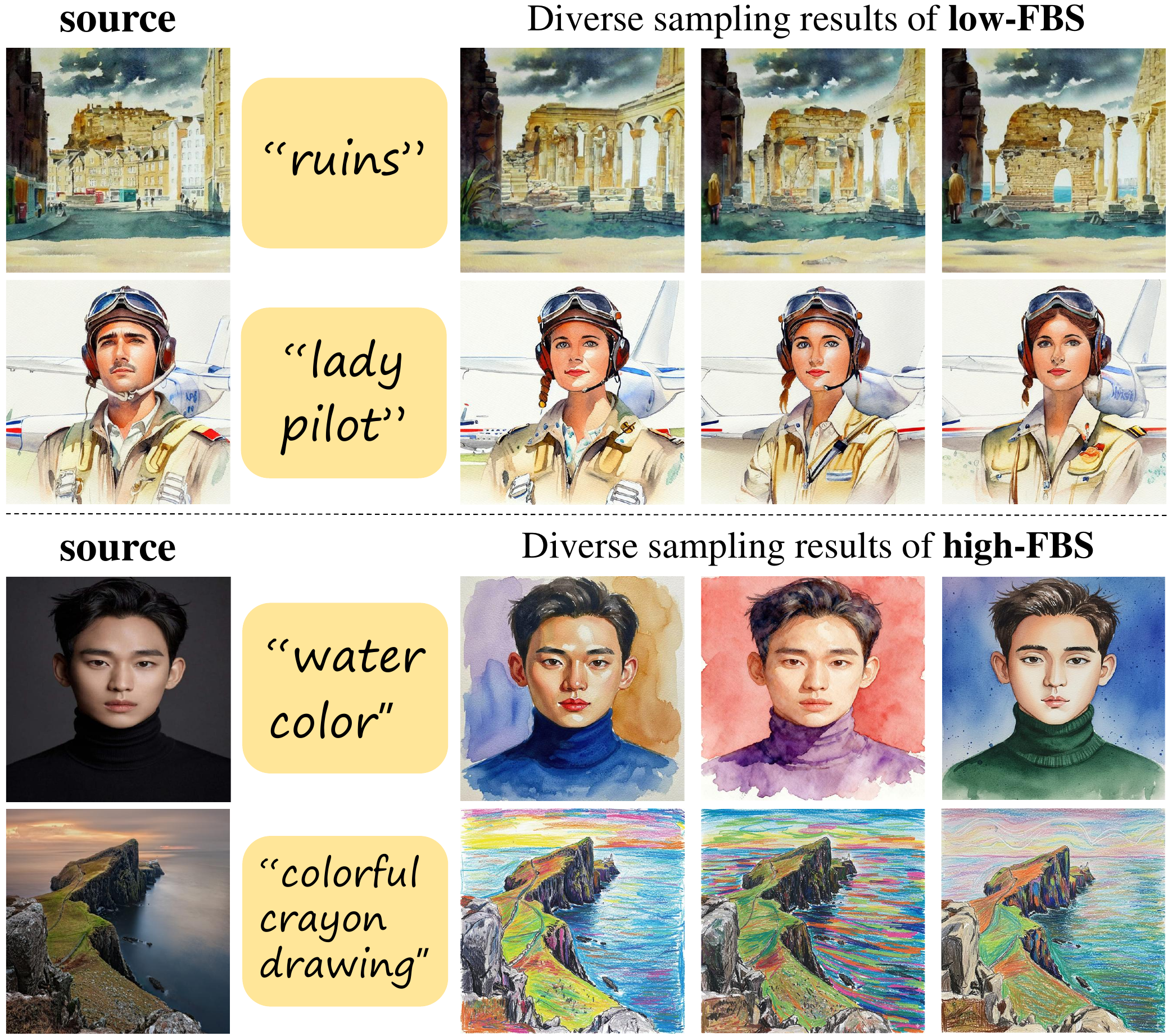}
		\caption{Our method features sampling diversity. FBSDiff++ generates random and diverse I2I results at each inference time. \textbf{Better viewed with zoom-in}.}
		\label{fig:diverse_results}
	\end{figure}
	
	The I2I correlations achieved by our method in different modes of frequency band substitution are more clearly demonstrated in Fig. \ref{fig:high_mid_low_compare}. When tested with the same source image and text prompt, result inherits the appearance (including style, hue, and structure) of the source image in low-FBS mode; maintains source image contours but exhibits style change in high-FBS mode; preserves pure image layout, allowing both style change and contour deviation in mid-FBS mode.
	
	One advantage of FBSDiff and FBSDiff++ over other text-driven I2I methods is sampling diversity, which is realized by randomly sampling $\tilde{z}_{T}$, the initial noise feature of the sampling trajectory, from the normal Gaussian distribution at each inference time. As Fig. \ref{fig:diverse_results} displays, FBSDiff++ allows to sample diverse I2I results based on a given pair of source image and text prompt, which differs from other inversion-based I2I methods that generate only a single result. 
	
	\begin{figure*}[t]
		\centering
		\includegraphics[width=0.99\linewidth]{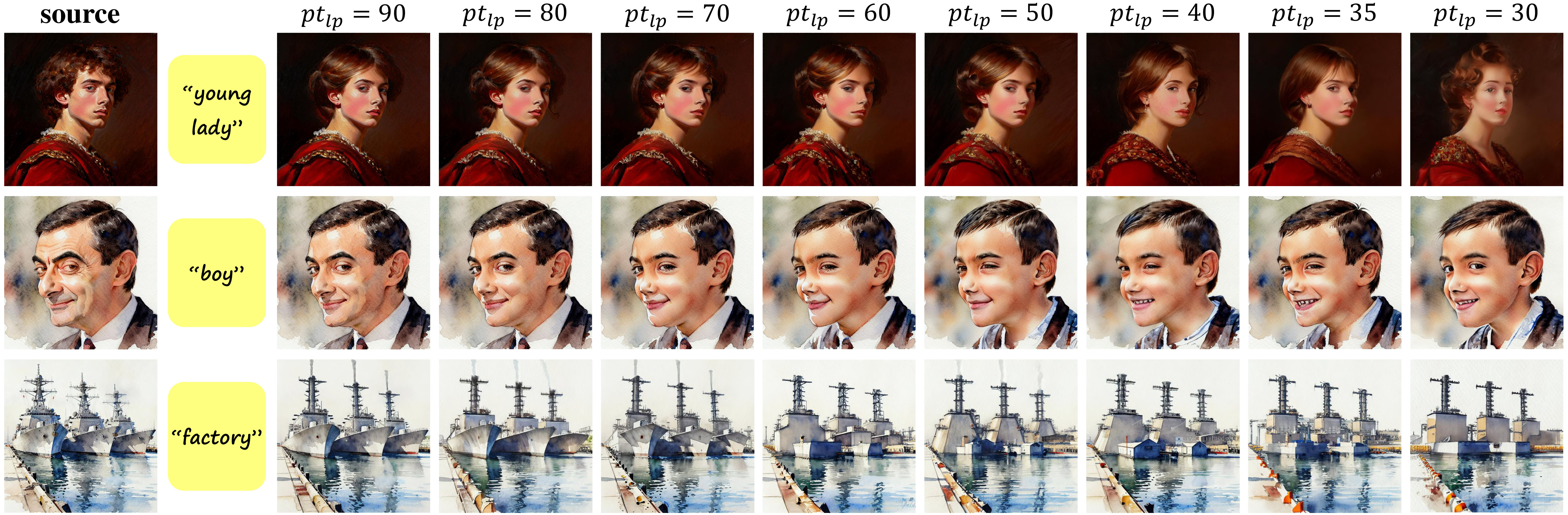}
		\caption{I2I appearance correlation control of FBSDiff++ achieved by tuning the low-FBS percentile threshold $pt_{lp}$. The I2I appearance consistency increases with the growing of $pt_{lp}$. Smaller $pt_{lp}$ leads to more variation of the translated image.}
		\label{fig:low-FBS_consistency_control}
	\end{figure*}
	
	\begin{figure*}[t]
		\centering
		\includegraphics[width=0.99\linewidth]{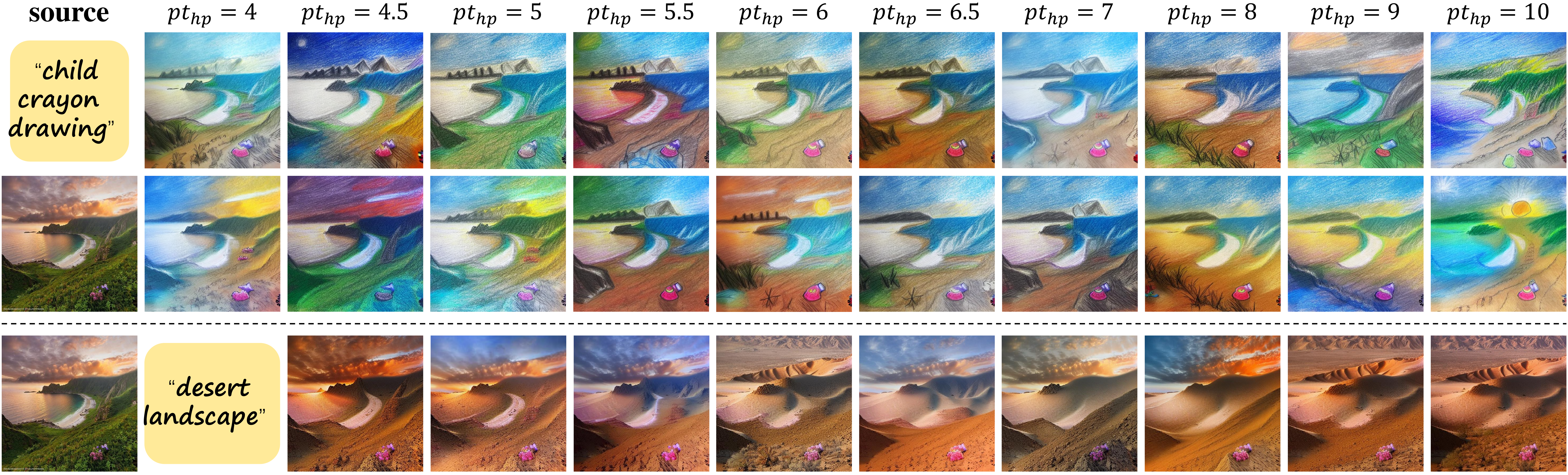}
		\caption{I2I contour correlation control of FBSDiff++ achieved by tuning the high-FBS percentile threshold $pt_{hp}$. The I2I contour consistency increases as $pt_{hp}$ declines. Larger $pt_{hp}$ leads to more contour deviation of the translated image.}
		\label{fig:high-FBS_consistency_control}
	\end{figure*}
	
	FBSDiff and FBSDiff++ also feature highly controllable I2I correlation intensity. As shown in Fig. \ref{fig:low-FBS_consistency_control}, for the application of derivative image generation realized by low-FBS, FBSDiff++ enables users to flexibly and continuously control the degree of I2I appearance consistency simply by tuning the low-pass percentile threshold $pt_{lp}$ in the normalized range of [0, 100]. Larger value of $pt_{lp}$ corresponds to wider bandwidth of the substituted frequency band such that more spectral components of $z_{t}$ are transplanted into $\tilde{z}_{t}$ during the sampling process, yielding translated images with a closer resemblance to the source image. Conversely, smaller value of $pt_{lp}$ corresponds to narrower bandwidth of the substituted frequency band, and accordingly leads to translated images with more visual variation. Such I2I correlation intensity control also applies to other modes of AdaFBS. Taking the image style translation task realized by high-FBS for example, FBSDiff++ enables continuous control over I2I contour consistency simply by tuning the high-pass percentile threshold $pt_{hp}$. In the mode of high-FBS, smaller value of $pt_{hp}$ corresponds to wider bandwidth of the substituted frequency band and thus yields the translated image with higher contour consistency to the source image. Conversely, an increase in $pt_{hp}$ leads to more contour deviation of the translated image as compared to the source image. Since contour consistency largely determines structure similarity between two images, the tuning of $pt_{hp}$ in high-FBS mode also indirectly controls I2I structure consistency for the image style translation task.
	
	We compare our methods with a comprehensive set of leading-edge and competitive baseline models, including Null-text Inversion (Null-text) \cite{mokady2023null}, pix2pix-zero (p2p-zero) \cite{parmar2023zero}, Plug-and-Play (PnP) \cite{tumanyan2023plug}, Prompt-tuning inversion (PT-Inv) \cite{dong2023prompt}, StyleDiffusion (StyleDiff) \cite{li2023stylediffusion}, Accelerated Iterative Diffusion Inversion (AIDI) \cite{pan2023effective}, Editable Noise Map Inversion (ENM-Inv) \cite{kang2025editable}, Negative-prompt inversion (Neg-Inv) \cite{miyake2025negative}, Renoise \cite{garibi2024renoise}, Source prompt disentangled inversion (SPDInv) \cite{li2024source}, Edit Friendly \cite{huberman2024edit}, Instructpix2pix (Ins-p2p) \cite{brooks2023instructpix2pix}, Sine \cite{zhang2023sine}, Imagic \cite{kawar2023imagic}, Masactrl \cite{cao2023masactrl}, IC-Edit \cite{zhang2025context}, GNRI \cite{samuel2024lightning}, and Swiftedit \cite{nguyen2025swiftedit}.
	
	\begin{figure*}[t]
		\centering
		\includegraphics[width=0.985\linewidth]{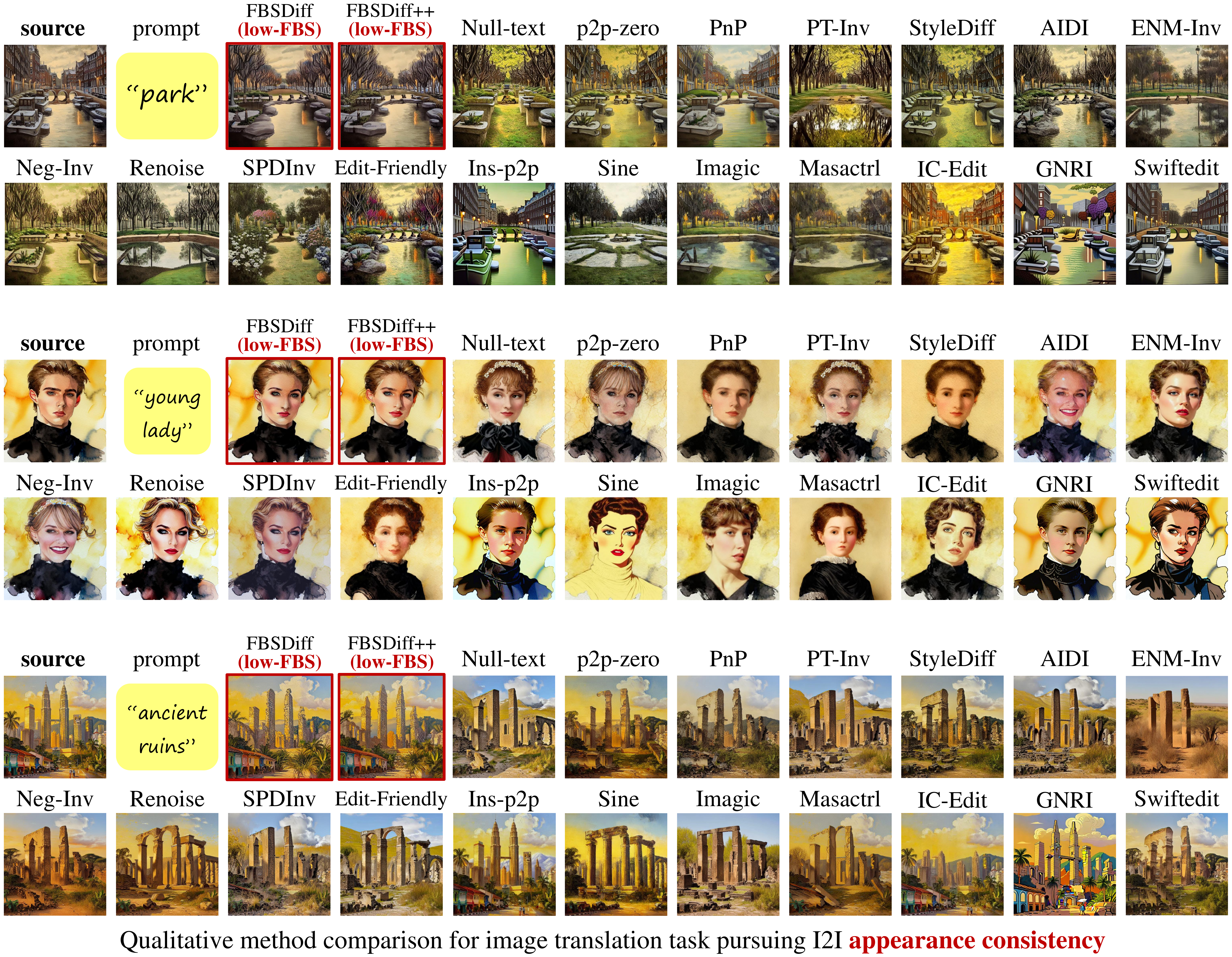}
		\caption{Qualitative method comparison for I2I task pursuing appearance consistency. In contrast to related methods, results of FBSDiff and FBSDiff++ in the mode of low-FBS more completely inherit source image style and structure without sacrificing text fidelity, which makes them more suitable to derivative image generation. \textbf{Better viewed with zoom-in}.}
		\label{fig:appearance_consistency_method_compare}
	\end{figure*}
	
	\begin{figure*}[t]
		\centering
		\includegraphics[width=0.985\linewidth]{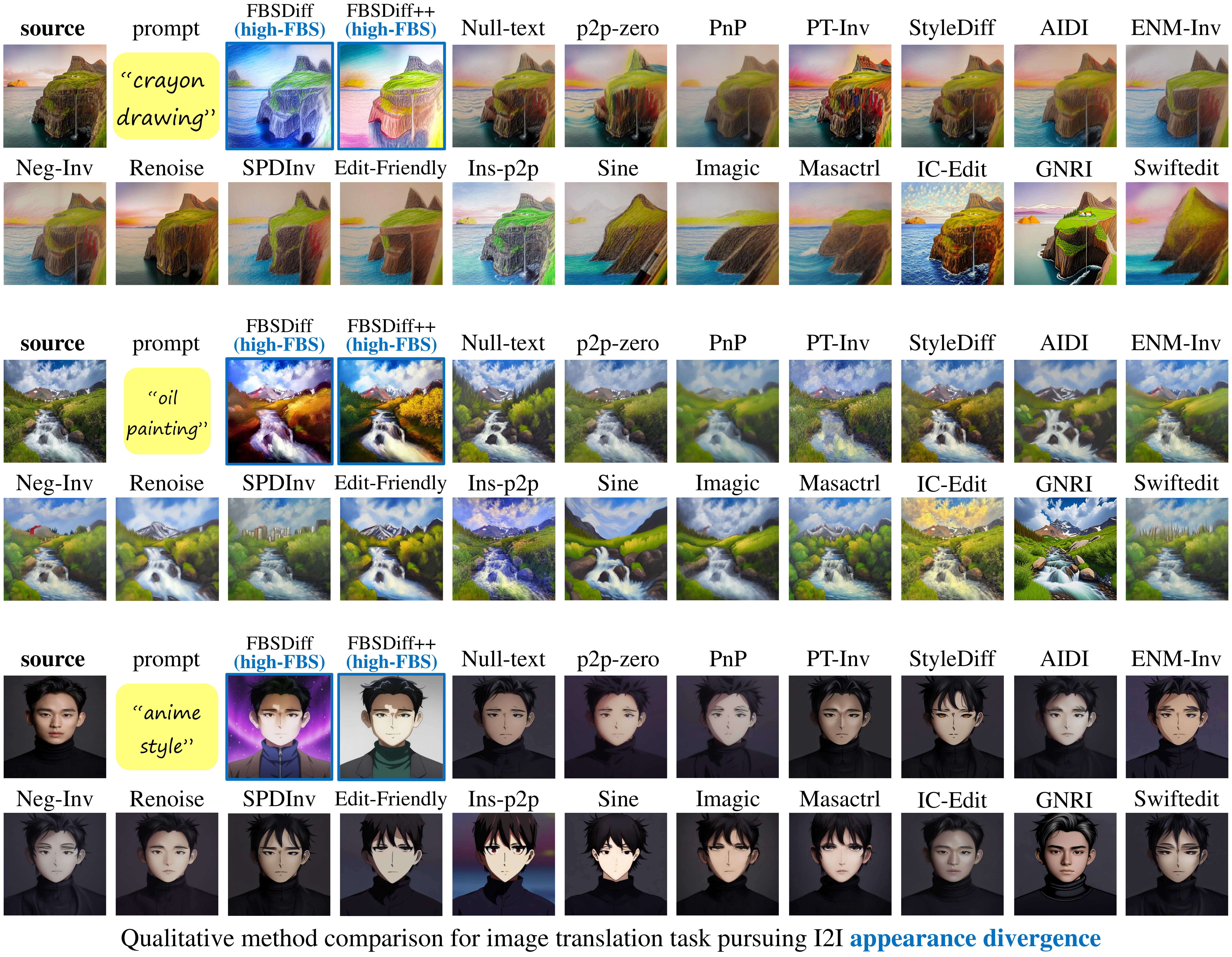}
		\caption{Qualitative method comparison for I2I task pursuing appearance divergence. In contrast to related methods, FBSDiff and FBSDiff++ in the mode of high-FBS are much more capable of extensively altering image appearance (e.g., color distribution), which makes them more suitable to text-driven image style translation. \textbf{Better viewed with zoom-in}.}
		\label{fig:appearance_divergence_method_compare}
	\end{figure*}
	
	Since our methods apply to different I2I applications with different modes of frequency band substitution, we correspondingly evaluate methods from different emphases. Fig. \ref{fig:appearance_consistency_method_compare} shows qualitative method comparison for the I2I application of derivative image generation, where appearance (including style, hue, structure) consistency between the source image and the translated image is pursued. Among the compared methods, Sine, Imagic, and Masactrl are relatively weak in structure preservation. Ins-p2p and IC-Edit achieve high structure consistency but tend to produce results that are not semantically faithful to the text prompt. Other baseline models perform well in structure similarity and text fidelity, but are not capable of precisely inheriting the style and hue of the source image. In contrast, results of FBSDiff and FBSDiff++ in the mode of low-FBS present noticeably higher appearance consistency, the translated image closely resembles the source image in structure, aesthetic style, texture, and hue, all while faithfully conforms to the text prompt in semantics, achieving satisfactory performance for derivative image generation. Fig. \ref{fig:appearance_divergence_method_compare} shows qualitative method comparison for the I2I application of image style translation where appearance alteration and diversity is favored, i.e., we pursue the goal of significantly changing the style of the source image and generating random and diverse stylization results. Given a text that indicates a specific style, our methods in high-FBS mode precisely transform the source image into the instructed style with the original image contours basically maintained. Results in Fig. \ref{fig:appearance_divergence_method_compare} demonstrate that FBSDiff and FBSDiff++ in the mode of high-FBS exhibit a distinct advantage in rerendering image appearance (including style and hue) compared to other baseline models. This is because in our methods, the low-frequency components of the source image related to image style and hue are explicitly filtered out in the mode of high-FBS, which effectively diminishes I2I appearance correlation and thus facilitates generating translated images with random, diverse, and significantly changed image appearance. By contrast, other related methods struggle to produce stylization results with large appearance divergence.
	
\begin{figure*}[t]
		\centering
		\includegraphics[width=0.99\linewidth]{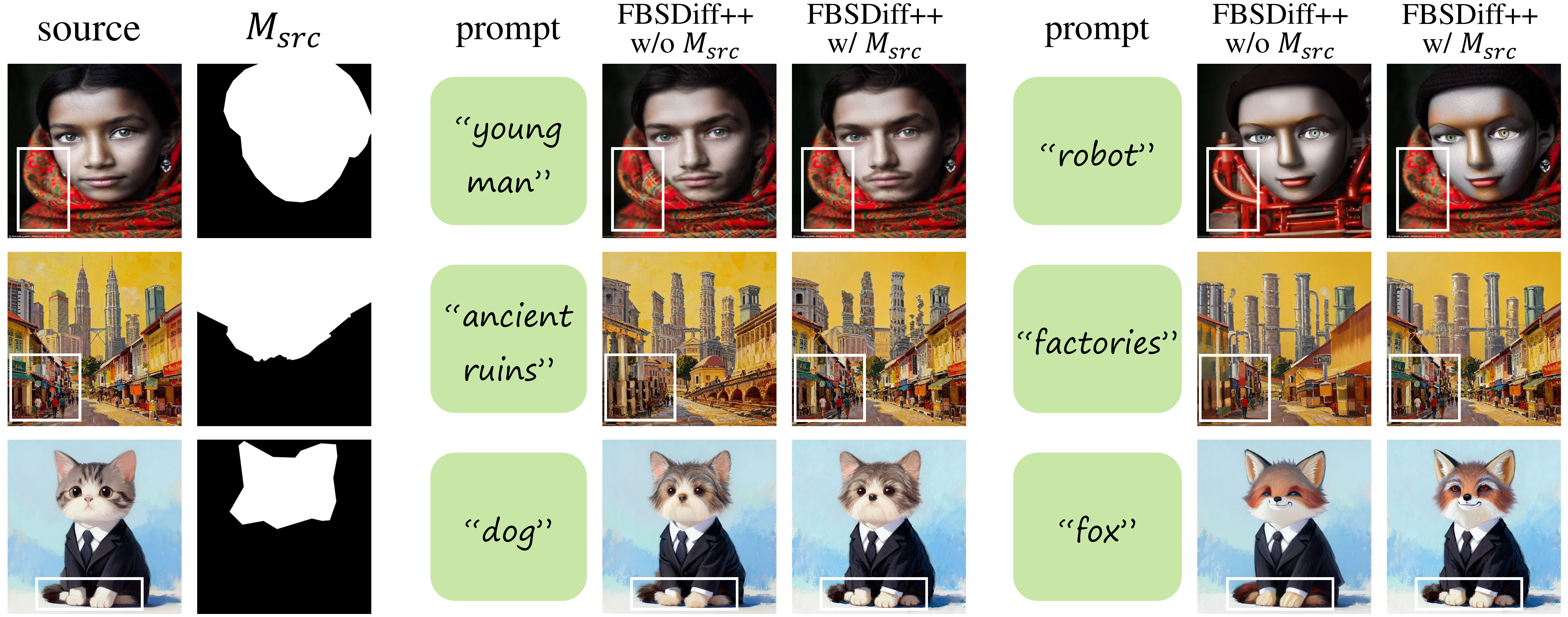}
		\caption{Demonstration of the extended functionality of FBSDiff++ for localized image manipulation. FBSDiff++ enables to locally manipulate the source image only in a specific region with a text prompt, where the manipulated area is indicated by a binary image mask $M_{src}$. Content of the translated image outside the selected area is strictly maintained as the same as the source image. The implementation is described in section \ref{extended}. \textbf{Better viewed with zoom-in}.}
		\label{fig:localized_image_manipulation}
	\end{figure*}

	Fig. \ref{fig:localized_image_manipulation} demonstrates the extended functionality of FBSDiff++ for localized image manipulation. FBSDiff++ allows to textually manipulate only a local region of the source image where the manipulated area is indicated by a binary mask $M_{src}$. Content of the source image outside the selected region of $M_{src}$ is intactly preserved. Taking the first row of Fig. \ref{fig:localized_image_manipulation} for example, when translating a source image of a young lady into a ``young man", introducing $M_{src}$ that indicates the head region results in the translated image with only the character's head semantically modified, while the clothing is completely unchanged (see the patterns on the clothing). Otherwise, normal FBSDiff++ without $M_{src}$ leads to the translated image with both the head and the clothing changed. Though FBSDiff++ is sufficient for high-quality text-driven I2I, the extended algorithm with $M_{src}$ endows the model with finer-grained control to locally manipulate an image.

	\begin{figure}[t]
		\centering
		\includegraphics[width=0.98\linewidth]{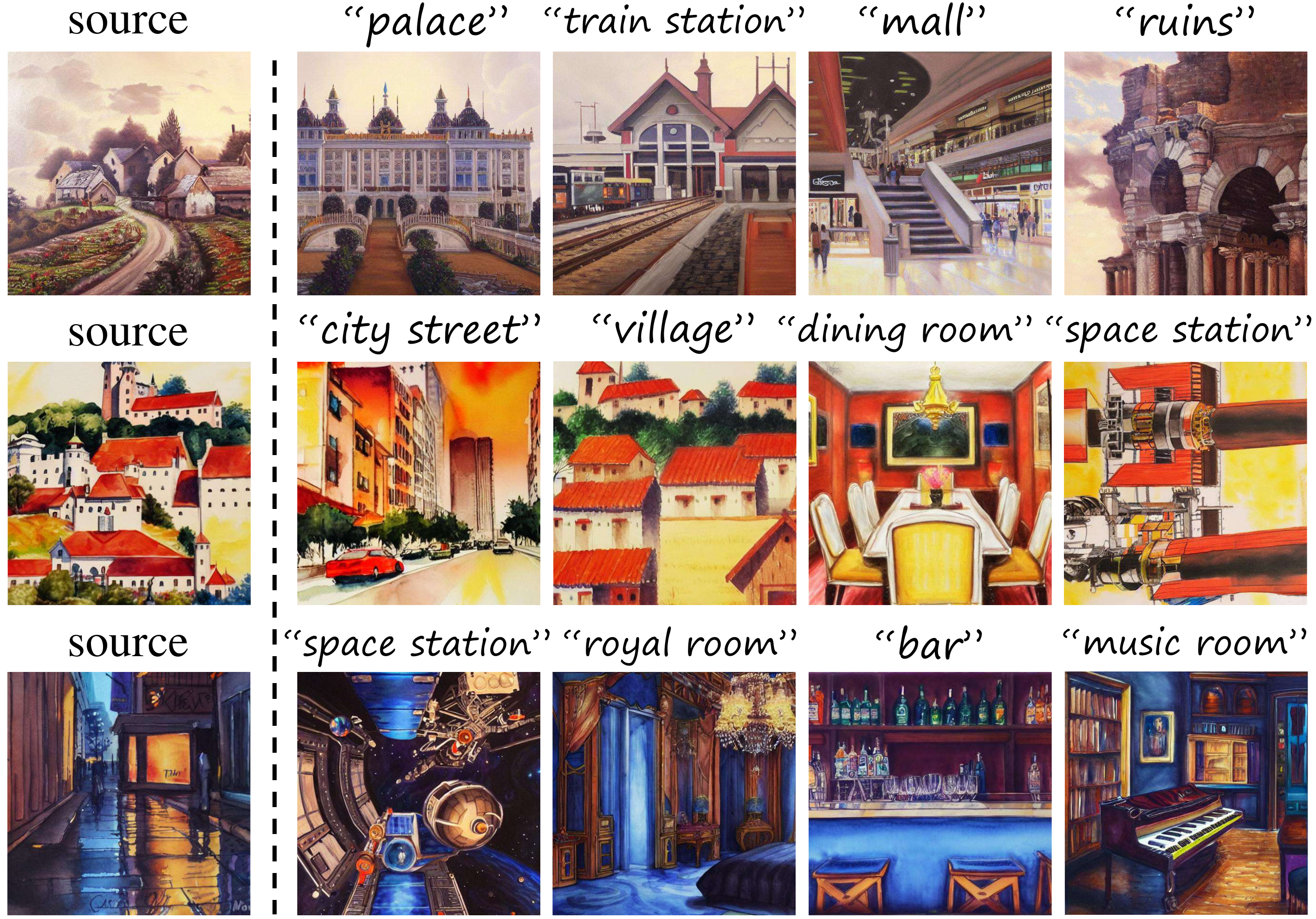}
		\caption{Demonstration of the extended functionality of FBSDiff++ for style-specific content creation. The translated image inherits only style distribution of the source image without structural correlation. The implementation is described in section \ref{extended}. \textbf{Better viewed with zoom-in}.}
		\label{fig:style_specific_content_creation}
	\end{figure} 
	
\begin{figure}[t]
		\centering
		\includegraphics[width=0.99\linewidth]{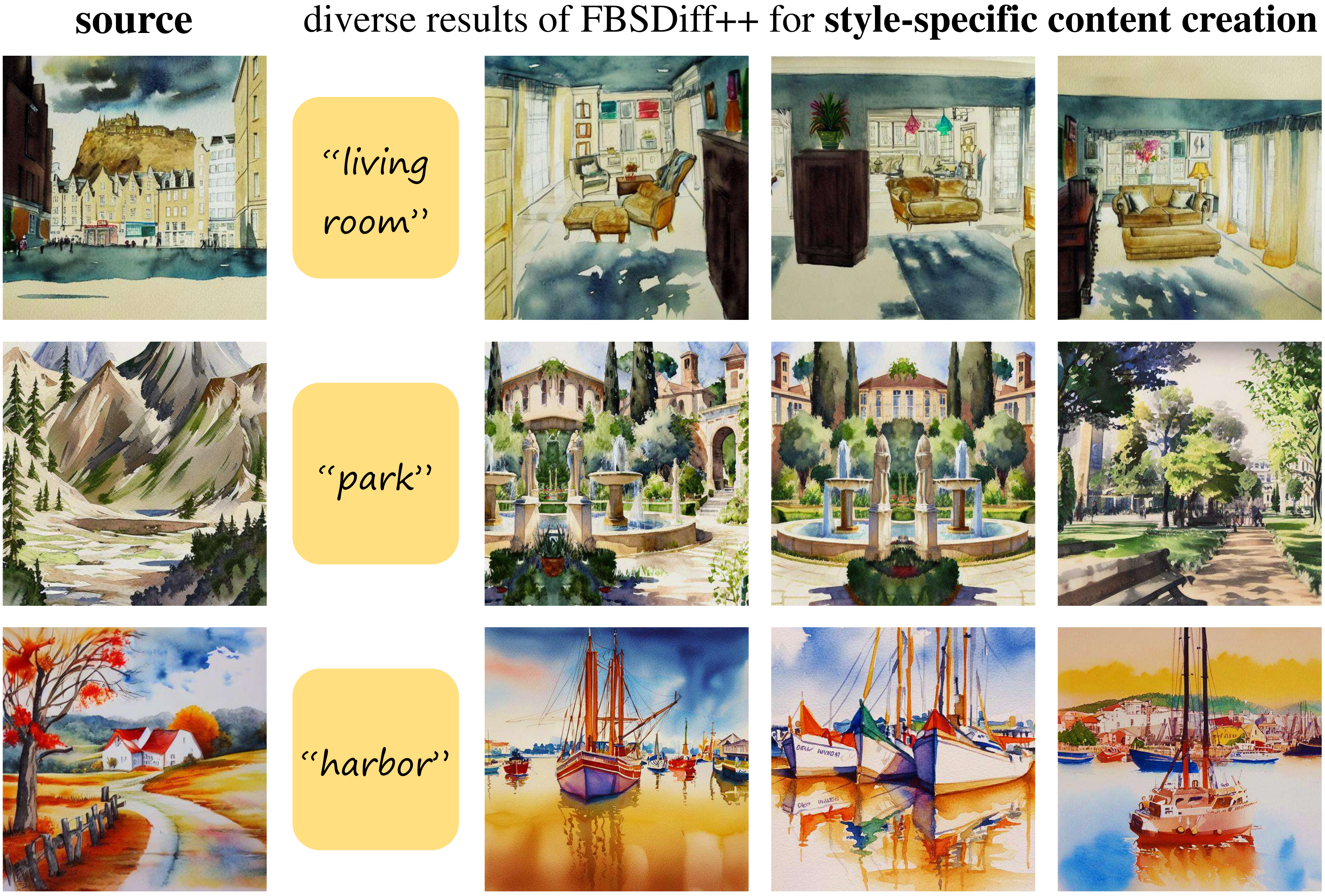}
		\caption{FBSDiff++ samples diverse I2I results for the extended functionality of style-specific content creation. For a given text prompt, all results correlate with the source image only in style, while the image structure is disentangled and random. \textbf{Better viewed with zoom-in}.}
		\label{fig:style_specific_content_creation_diverse}
	\end{figure}

	Fig. \ref{fig:style_specific_content_creation} demonstrates the extended functionality of FBSDiff++ for style-specific content creation, which is realized by embedding the STP module into the normal FBSDiff++ framework in the mode of low-FBS. This functionality enables to generate stylistically consistent but structurally disentangled I2I results, i.e., the translated images derive from the source image solely in terms of style. Fig. \ref{fig:style_specific_content_creation_diverse} further shows that, given a fixed pair of source image and text prompt, the extended application for style-specific content creation allows to sample structurally random and diverse I2I results that share the same style as the source image. Such randomness and diversity are partly due to the randomly initialized noise feature $\tilde{z}_{T}$ at the beginning of the sampling trajectory, and partly due to the randomly sampled spatial transformation hyper-parameters in the STP module at each inference time. 
	
	\begin{figure*}[t]
		\centering
		\includegraphics[width=0.98\linewidth]{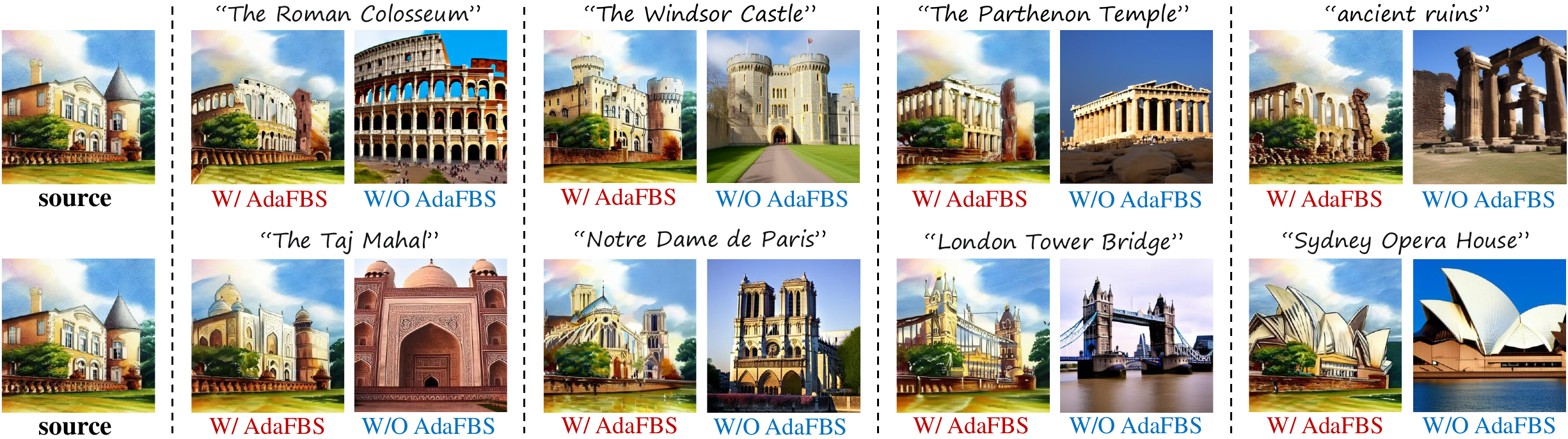}
		\caption{Ablation study of FBSDiff++ w.r.t. the AdaFBS module. Taking the \textbf{low-FBS} mode for example, results w/ AdaFBS exhibit I2I appearance consistency, while results w/o AdaFBS have no visual correlation with the source image.}
		\label{fig:FBS_ablation}
	\end{figure*}
	
	\begin{figure*}[t]
		\centering
		\includegraphics[width=0.985\linewidth]{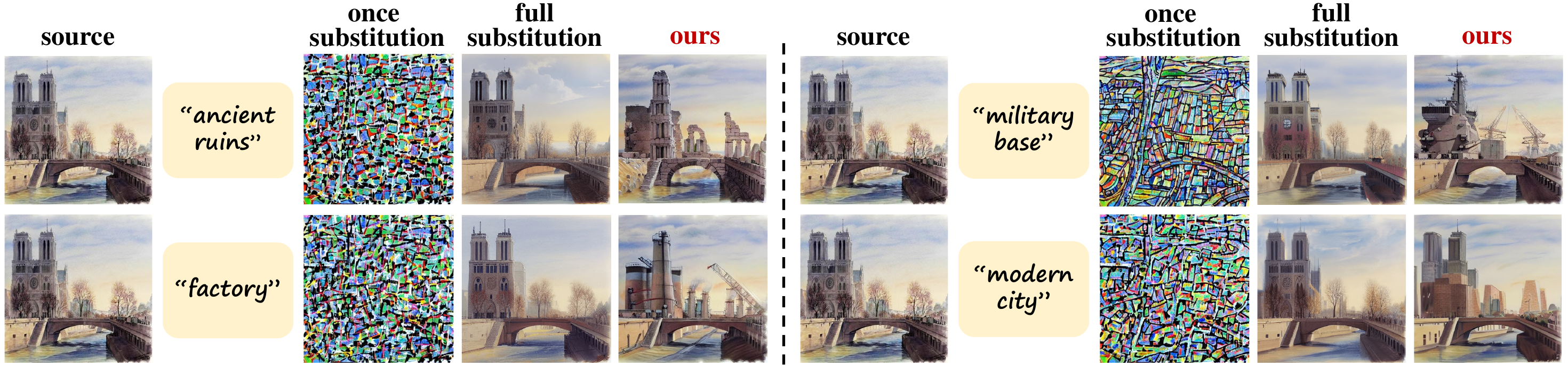}
		\caption{Ablation study w.r.t. different manners of frequency band substitution. \textbf{Better viewed with zoom-in}. }
		\label{fig:substitute_ablation}
	\end{figure*}
	
	\begin{figure*}[t]
		\centering
		\includegraphics[width=0.985\linewidth]{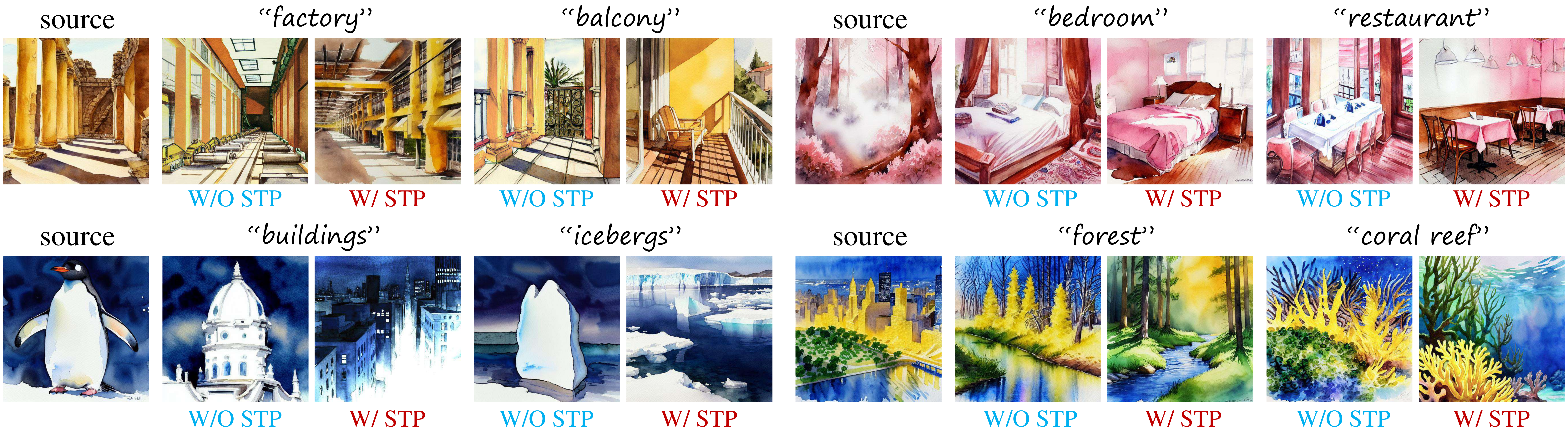}
		\caption{Ablation study of the Spatial Transformation Pool (STP) for the extended functionality of style-specific content creation. Results w/o STP correlate with the source image both in image style and image structure, while results w/ STP disentangle image structure and inherit only style distribution of the source image. \textbf{Better viewed with zoom-in}. }
		\label{fig:STP_ablation}
	\end{figure*}
	
	To verify the rationality and validity of our methods, we conduct ablation studies w.r.t. the kernel ingredient of our approach, different forms of frequency band substitution, and the pivotal module for the extended functionality. 
	
	Ablation study w.r.t. the AdaFBS module, the critical component of FBSDiff++, is illustrated in Fig. \ref{fig:FBS_ablation}. Taking the low-FBS mode for example, the framework w/o AdaFBS leads to results with no correlation to the source image, violating the basic premise of I2I translation, while the model w/ AdaFBS produces results that are both highly consistent to the source image in appearance and faithful to the text prompt in semantics. 
	
	We explore and compare our method with other potential designs of frequency band substitution, including substituting frequency band only once at $\lambda T$ time step rather than along all the first $\lambda T$ steps of the sampling trajectory (which we denote \textbf{once substitution}), and substituting the full DCT spectrum rather than a partial sub-band of it (which we refer to as \textbf{full substitution}). As shown in Fig. \ref{fig:substitute_ablation} (taking the low-FBS mode for example), \textbf{once substitution} at only an intermediate time step produces completely corrupted results, suggesting that per-step feature calibration at the early stage of the sampling trajectory is of crucial necessity for smooth feature fusion. Results of \textbf{full substitution} suffer from weak text fidelity (i.e., not faithful to the text semantics). This is because substituting full DCT spectrum transfers excessive visual information of the source image, such that the translated image is too consistent with the source image to semantically adhere to the target text.
	
	Ablation study in Fig. \ref{fig:STP_ablation} highlights the key role of the STP module for the extended functionality of style-specific content creation. Results w/o STP module are structurally consistent with the source image, violating the goal of deriving \textbf{only style} from the source image. In contrast, results w/ STP module are structurally disentangled from the source image, indicating the effectiveness of STP in randomly shuffling image structure. 
	
	\begin{table*}[t]\large
		\begin{center}
			\caption{Evaluations on LAION-Mini for both I2I tasks emphasizing appearance consistency and appearance divergence.}
			\label{tab:metrics}
			\resizebox{\linewidth}{!}{
				\begin{threeparttable}
					\begin{tabular}{|c|ccccc|cccc|}
						\hline
						{\textbf{Emphasis}} & \multicolumn{5}{|c|}{\textbf{Pursue appearance consistency}} & \multicolumn{4}{|c|}{\textbf{Pursue appearance divergence}} \\
						\hline
						\diagbox[]{\textbf{Methods}}{\textbf{Metrics}} & \makecell[c]{Structure\\Similarity($\uparrow$)} & \makecell[c]{LPIPS($\downarrow$)} & \makecell[c]{AdaIN Style\\ Loss($\downarrow$)} & \makecell[c]{CLIP\\Similarity($\uparrow$)} & \makecell[c]{Aesthetic\\Score($\uparrow$)} &  \makecell[c]{Structure\\Similarity($\uparrow$)} & \makecell[c]{AdaIN Style\\ Loss($\uparrow$)} & \makecell[c]{CLIP\\Similarity($\uparrow$)} & \makecell[c]{Aesthetic\\Score($\uparrow$)}\\
						
						\hline
						{Null-text \cite{mokady2023null}} & 0.947 & 0.249 & \textcolor{blue}{18.025} & 0.278 & 6.501 & 0.950 & 23.786 & 0.271 & 6.404 \\
						
						{p2p-zero \cite{parmar2023zero}} & 0.950 & \textcolor{blue}{0.245} & \textcolor{blue}{17.244} & 0.264 & 6.476 & 0.955 & 21.403 & 0.260 & 6.337 \\
						
						{PnP \cite{tumanyan2023plug}} & 0.955 & 0.268 & 21.003 & \textcolor{red}{0.286} & 6.585 & 0.958 & 27.625 & 0.276 & 6.456 \\
						
						{PT-Inv \cite{dong2023prompt}} & 0.946 & 0.248 & 21.055 & 0.273 & 6.487 & 0.949 & 25.128 & 0.267 & 6.320 \\
						
						{StyleDiff \cite{li2023stylediffusion}} & 0.944 & 0.253 & 22.245 & 0.270 & 6.479 & 0.946 & 25.837 & 0.265 & 6.309 \\
						
						{AIDI \cite{pan2023effective}} & 0.953 & 0.251 & 18.683 & 0.277 & 6.496 & 0.954 & 23.667 & 0.273 & 6.286 \\
						
						{ENM-Inv \cite{kang2025editable}} & 0.948 & 0.269 & 23.108 & 0.283 & 6.532 & 0.945 & 24.524 & 0.277 & 6.411 \\
						
						{Neg-Inv \cite{miyake2025negative}} & 0.945 & 0.252 & 18.975 & 0.278 & 6.488 & 0.952 & 23.584 & 0.272 & 6.367 \\
						
						{Renoise \cite{garibi2024renoise}} & 0.938 & 0.270 & 22.685 & 0.281 & 6.516 & 0.953 & 24.450 & 0.268 & 6.393 \\
						
						{SPDInv \cite{li2024source}} & 0.949 & 0.264 & 22.363 & 0.277 & 6.489 & 0.942 & 25.267 & 0.278 & 6.352 \\
						
						{Edit-Friendly \cite{huberman2024edit}} & 0.940 & 0.265 & 23.457 & 0.279 & 6.510 & 0.940 & 27.479 & \textcolor{blue}{0.280} & 6.409 \\
						
						{Ins-P2P \cite{brooks2023instructpix2pix}} & \textcolor{blue}{0.959} & 0.263 & 23.106 & 0.261 & 6.274 & 0.962 & \textcolor{blue}{31.242} & 0.266 & 6.240 \\
						
						{Sine \cite{zhang2023sine}} & 0.934 & 0.274 & 24.184 & \textcolor{blue}{0.285} & 6.339 & 0.937 & 28.336 & \textcolor{red}{0.283} & 6.175 \\
						
						{Imagic \cite{kawar2023imagic}} & 0.932 & 0.271 & 21.388 & 0.282 & 6.476 & 0.935 & 27.724 & 0.279 & 6.213 \\
						
						{Masactrl \cite{cao2023masactrl}} & 0.937 & 0.257 & 19.696 & 0.280 & 6.508 & 0.941 & 26.250 & 0.274 & 6.197 \\
						
						{IC-Edit \cite{zhang2025context}} & 0.954 & 0.254 & 18.886 & 0.268 & \textcolor{red}{6.593} & \textcolor{red}{0.968} & \textcolor{blue}{32.365} & 0.266 & \textcolor{red}{6.470} \\
						
						{GNRI \cite{samuel2024lightning}} & \textcolor{blue}{0.958} & 0.250 & 20.869 & 0.265 & \textcolor{blue}{6.589} & \textcolor{blue}{0.965} & 29.587 & 0.262 & \textcolor{blue}{6.468} \\
						
						{Swiftedit \cite{nguyen2025swiftedit}} & 0.952 & \textcolor{blue}{0.247} & 19.574 & 0.274 & 6.523 & 0.939 & 26.728 & 0.270 & 6.228 \\
						
						{\textbf{FBSDiff (ours)}} & \textcolor{blue}{0.963} & \textcolor{blue}{0.241} & \textcolor{blue}{15.758} & \textcolor{blue}{0.285} & \textcolor{blue}{6.586} & \textcolor{blue}{0.965} & \textcolor{red}{34.323} & \textcolor{blue}{0.281} & \textcolor{blue}{6.462} \\
						
						{\textbf{FBSDiff++ (ours)}} & \textcolor{red}{0.965} & \textcolor{red}{0.240} & \textcolor{red}{15.577} & \textcolor{blue}{0.284} & \textcolor{blue}{6.587} & \textcolor{blue}{0.966} & \textcolor{blue}{34.166} & \textcolor{blue}{0.280} & \textcolor{blue}{6.465} \\
						
						\hline
					\end{tabular}
					\begin{tablenotes}
						\item The \textcolor{red}{red font} denotes the best result, the \textcolor{blue}{blue font} denotes the next three runners-up.
					\end{tablenotes}
				\end{threeparttable}
			}
		\end{center}
	\end{table*}
	
	\begin{figure*}[t]
		\centering
		\includegraphics[width=0.99\linewidth]{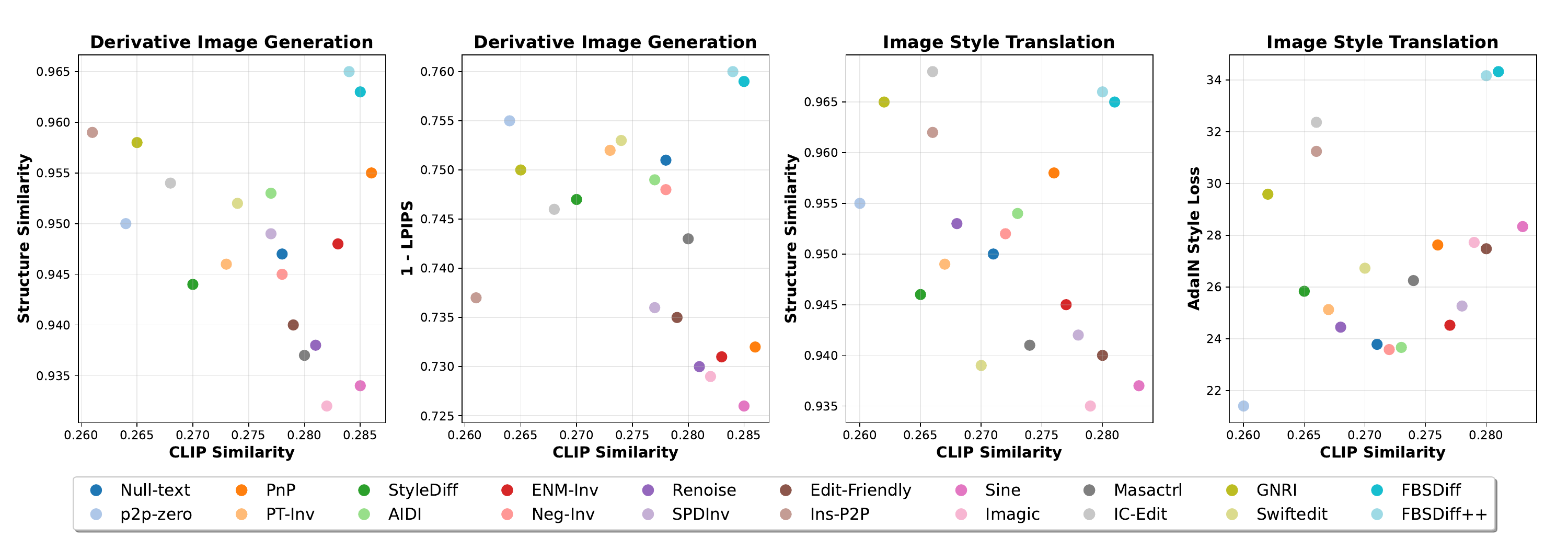}
		\caption{Scatter plot comparison of different methods on two I2I tasks: derivative image generation (left two panels) and image style translation (right two panels). For all scatter plots, a location closer to the top-right corner is more desirable.}
		\label{fig:quantitative_eval}
	\end{figure*}
	
	\begin{table*}[t]
		\centering
		\footnotesize
		\renewcommand{\arraystretch}{0.9}
		\setlength{\tabcolsep}{8pt} 
		\caption{Evaluation of FBSDiff++ in different modes of AdaFBS with varying DCT filtering percentile thresholds.}
		\label{tab:performance_with_varying_thresholds}
		\begin{tabularx}{0.99\linewidth}{@{} l *{9}{>{\centering\arraybackslash}X} @{}} 
			\toprule
			\multicolumn{10}{>{\centering}p{\dimexpr\linewidth-2\tabcolsep\relax}}{
				Low-FBS Mode for Derivative Image Generation} \\
			\midrule
			$pt_{lp}$ & 90 & 80 & 70 & \textbf{60} & 50 & 40 & 30 & 20 & 10 \\
			\midrule
			Structure Similarity ($\uparrow$) & 0.988 & 0.983 & 0.975 & \textbf{0.965} & 0.962 & 0.959 & 0.954 & 0.951 & 0.949 \\
			LPIPS ($\downarrow$) & 0.145 & 0.197 & 0.221 & \textbf{0.240} & 0.268 & 0.301 & 0.343 & 0.362 & 0.384 \\
			CLIP Similarity ($\uparrow$) & 0.226 & 0.253 & 0.271 & \textbf{0.284} & 0.291 & 0.299 & 0.310 & 0.315 & 0.317 \\
			\midrule[0.8pt]
			\multicolumn{10}{>{\centering}p{\dimexpr\linewidth-2\tabcolsep\relax}}{
				High-FBS Mode for Image Style Translation} \\
			\midrule
			$pt_{hp}$ & 3.5 & 4.0 & 4.5 & \textbf{5.0} & 5.5 & 6.0 & 7.0 & 8.0 & 9.0 \\
			\midrule
			Structure Similarity ($\uparrow$) & 0.990 & 0.987 & 0.978 & \textbf{0.966} & 0.963 & 0.959 & 0.956 & 0.953 & 0.950 \\
			Style Loss ($\uparrow$) & 26.447 & 28.656 & 31.083 & \textbf{34.166} & 35.524 & 36.379 & 37.105 & 37.628 & 37.896 \\
			CLIP Similarity ($\uparrow$) & 0.247 & 0.261 & 0.272 & \textbf{0.280} & 0.286 & 0.293 & 0.299 & 0.304 & 0.308 \\
			\bottomrule
		\end{tabularx}
	\end{table*}
	
	\begin{table*}[!htbp]
		\centering
		\footnotesize
		\setlength{\tabcolsep}{0pt} 
		\renewcommand{\arraystretch}{0.9}
		\caption{Comparison between FBSDiff and FBSDiff++ with varying inversion steps (averaged over 20 samples).}
		\label{tab:FBSDiff_FBSDiff++_inversion_steps}
		\begin{tabular*}{0.99\linewidth}{@{\extracolsep{\fill}} l *{5}{c} *{5}{c} @{}}
			\toprule
			\multicolumn{1}{c}{} & \multicolumn{5}{c}{\textbf{FBSDiff}} & \multicolumn{5}{c}{\textbf{FBSDiff++}} \\
			\cmidrule(lr){2-6} \cmidrule(l){7-11}
			\textbf{Inversion Steps} & 50 & 100 & 200 & 500 & 1000 & 50 & 100 & 200 & 500 & 1000 \\
			\midrule
			Structure Similarity ($\uparrow$) & 0.942 & 0.944 & 0.947 & 0.956 & 0.981 & 0.981 & 0.982 & 0.982 & 0.982 & 0.983 \\
			LPIPS ($\downarrow$) & 0.596 & 0.511 & 0.473 & 0.456 & 0.237 & 0.238 & 0.238 & 0.237 & 0.237 & 0.236 \\
			CLIP Similarity ($\uparrow$) & 0.264 & 0.268 & 0.271 & 0.276 & 0.282 & 0.282 & 0.282 & 0.282 & 0.282 & 0.282 \\
			\bottomrule
		\end{tabular*}
	\end{table*}
	
	\begin{table*}[!htbp]
		\centering
		\footnotesize
		\setlength{\tabcolsep}{0pt} 
		\renewcommand{\arraystretch}{0.9}
		\caption{Ccomparison between FBSDiff and FBSDiff++ with varying source image size (averaged over 20 samples).}
		\label{tab:FBSDiff_FBSDiff++_resolution}
		\begin{tabular*}{0.99\linewidth}{@{\extracolsep{\fill}} l *{4}{c} *{4}{c} @{}}
			\toprule
			\multicolumn{1}{c}{} & \multicolumn{4}{c}{\textbf{FBSDiff}} & \multicolumn{4}{c}{\textbf{FBSDiff++}} \\
			\cmidrule(lr){2-5} \cmidrule(l){6-9}
			\textbf{Image Resolution} & 512$\times$512 & 512$\times$1024 & 1024$\times$512 & 2048$\times$2048 & 512$\times$512 & 512$\times$1024 & 1024$\times$512 & 2048$\times$2048 \\
			\midrule
			Structure Similarity ($\uparrow$) & 0.970 & 0.956 & 0.953 & 0.942 & 0.972 & 0.967 & 0.969 & 0.964 \\
			LPIPS ($\downarrow$) & 0.241 & 0.379 & 0.387 & 0.451 & 0.239 & 0.245 & 0.242 & 0.244 \\
			CLIP Similarity ($\uparrow$) & 0.293 & 0.252 & 0.256 & 0.290 & 0.291 & 0.288 & 0.293 & 0.287 \\
			\bottomrule
		\end{tabular*}
	\end{table*}

	\subsection{Quantitative evaluations}
	For quantitative method evaluations, we separately evaluate methods on the text-driven I2I task pursuing appearance consistency and the task pursuing appearance divergence. For the former task, we assess models' appearance preservation capability (including both structure consistency and style consistency) by measuring Structure Similarity ($\uparrow$), Perceptual Similarity ($\uparrow$), and Style Distance ($\downarrow$) between the source image and the translated image. For the latter task, we assess models' structure preservation and appearance (including style and hue) alteration abilities by measuring Structure Similarity ($\uparrow$) and Style Distance ($\uparrow$) between the I2I pairs. We use the DINO-ViT self-similarity distance \cite{chung2022improving} as the metric for Structure Distance, and define Structure Similarity as 1 - Structure Distance. We use LPIPS \cite{zhang2018unreasonable} metric to measure Perceptual Similarity, and use AdaIN style loss \cite{huang2017arbitrary} to measure Style Distance. Besides, CLIP Similarity ($\uparrow$) \cite{radford2021learning} is used to assess text fidelity, i.e., semantic faithfulness of the translated image to the text prompt. Finally, the Aesthetic Score ($\uparrow$) predicted by the LAION Aesthetics Predictor V2 model is used to evaluate the generation quality of the translated image. 
	
	The quantitative evaluations are conducted on both the appearance-consistency-oriented and appearance-divergence-oriented I2I tasks. For the evaluation dataset, we build LAION-Mini which contains 800 source images for each of the two tasks, with each image assigned two manually designed text prompts (1600 image-text pairs for each task). The images in our LAION-Mini dataset are sourced from both the web and a curated selection from the LAION Aesthetics 6.5+ dataset to ensure high-level aesthetic quality, and comprise both real-world photographs and artistic paintings. The designed text prompts focus on semantic content translation for the appearance-consistency-oriented task, and emphasize image style transformation for the appearance-divergence-oriented one. All source images in LAION-Mini are cropped to $512\times512$.
	
	To compare with other baseline methods, our FBSDiff and FBSDiff++ are evaluated under low-FBS for the task pursuing appearance consistency, and are evaluated under high-FBS for the task favoring appearance divergence. The average values of all the quantitative metrics evaluated on the LAION-Mini dataset are reported in Tab. \ref{tab:metrics}. Our FBSDiff and FBSDiff++ consistently secure a position within the top 4 methods across all evaluated metrics, indicating superiority of our methods in adapting to I2I tasks with different emphases. The competitive results in CLIP Similarity and Aesthetic Score also reflect high-level text fidelity and visual quality achieved by our methods. Fig. \ref{fig:quantitative_eval} leverages scatter plots to visually highlight advantages of our methods. For application of derivative image generation, FBSDiff and FBSDiff++ achieve more favorable trade-off between CLIP Similarity ($
	\uparrow$) and Structure Similarity ($\uparrow$) (most top-right positions in the first plot), as well as between CLIP Similarity ($
	\uparrow$) and LPIPS ($\downarrow$) (most top-right positions in the second plot) in the mode of low-FBS, indicating a superior balance achieved by our methods between text fidelity and I2I appearance consistency. Likewise, the third and fourth plots illustrate that for the application of image style translation that favors style alteration, FBSDiff and FBSDiff++ are more capable of producing substantial I2I style discrepancies, all while achieving superior trade-off between text fidelity and structure consistency.
	
	We have explored influence of the AdaFBS percentile thresholds to the performance of FBSDiff++, the average metric values evaluated on the LAION-Mini dataset are reported in Tab. \ref{tab:performance_with_varying_thresholds}. For derivative image generation realized with low-FBS, the decline of $pt_{lp}$ leads to weakened I2I structure consistency and perceptual similarity, while the text fidelity is accordingly improved. For image style translation realized with high-FBS, the increase of $pt_{hp}$ brings about improved text fidelity at the expense of degraded I2I structure consistency and style divergence. These numerical results are basically in line with the visual results demonstrated in Fig. \ref{fig:low-FBS_consistency_control} and Fig. \ref{fig:high-FBS_consistency_control}.
	
	To verify the substantial improvement in inference efficiency of FBSDiff++ over FBSDiff quantitatively, we evaluate the two methods over 20 samples selected from LAION-Mini under different inversion steps on the task of derivative image generation. The average evaluation results reported in Tab. \ref{tab:FBSDiff_FBSDiff++_inversion_steps} show that performance of FBSDiff++ is not sensitive to the number of inversion steps. This is due to that FBSDiff++ bypasses source image reconstruction process and thus places no demand on inversion accuracy. In contrast, FBSDiff suffers from severe performance degradation as the number of inversion steps decreases. FBSDiff++ with only 50 inversion steps can perform on par with FBSDiff with 1000 inversion steps.
	
	\begin{table}[t]
		\centering
		\footnotesize
		\caption{Inference speed comparison across different inversion-based methods.}
		\label{tab:inference_speed_comparison}
		\begin{tabularx}{0.99\linewidth}{l|c|c|c}
			\toprule
			\textbf{Method} & \textbf{Inversion} & \textbf{Sampling} & \textbf{Total} \\
			\midrule
			Null-text \cite{mokady2023null}     & 40.9s & 28.7s & 69.6s \\
			p2p-zero  \cite{parmar2023zero}     & 44.7s & 12.8s & 57.5s\\
			PnP \cite{tumanyan2023plug}         & 87.6s & 15.0s & 102.6s\\
			PT-Inv \cite{dong2023prompt}        & 42.4s & 29.9s & 72.3s\\
			StyleDiff \cite{li2023stylediffusion}     & 40.3s & 38.0s & 78.3s \\
			\textbf{FBSDiff} \cite{gao2024fbsdiff}        & 77.2s & 8.0s & 85.2s\\
			\textbf{FBSDiff++}      & \textcolor{red}{3.5s} & \textcolor{red}{6.1s} & \textcolor{red}{9.6s} \\
			\bottomrule
		\end{tabularx}
	\end{table}

	\begin{table}[t]
		\centering
		\footnotesize
		\caption{Quantitative ablation study w.r.t. the STP module for the task of style-specific content creation.}
		\label{tab:stp_ablation}
		\begin{tabularx}{0.99\linewidth}{l|c|c}
			\toprule
			\textbf{Metric} & \textbf{w/o STP} & \textbf{w/ STP}  \\
			\midrule
			Structure Similarity ($\downarrow$) & 0.973 & 0.916  \\
			Style Distance ($\downarrow$)      & 16.14 & 16.30 \\
			CLIP Similarity ($\uparrow$)      & 0.288 & 0.294 \\
			\bottomrule
		\end{tabularx}
	\end{table}
	
	Results in Tab. \ref{tab:FBSDiff_FBSDiff++_resolution} illustrate that FBSDiff++ is significantly more robust to the variations in the source image size than FBSDiff, benefiting from the improvements in adaptive frequency band substitution. Compared with FBSDiff, FBSDiff++ solves the issue of severe drop in CLIP Smilarity for non-square source images by adopting uniform-strength DCT filtering, and avoids degraded perceptual consistency (surge in LPIPS metric) for higher-resolution source images by employing percentile-based DCT thresholds. 
	
	Tab. \ref{tab:inference_speed_comparison} presents inference speed comparison among representative inversion-based text-driven I2I methods. Null-text \cite{mokady2023null}, PT-Inv \cite{dong2023prompt}, and StyleDiff \cite{li2023stylediffusion} entail relatively long sampling time due to the required online optimization process. FBSDiff features fast sampling but its inversion time remains relatively long. FBSDiff++ reduces the inversion steps from 1000 (in FBSDiff) to 50, cutting the inversion time from 77.2s to merely 3.5s, dramatically boosting inference speed. Moreover, by omitting the reconstruction trajectory, FBSDiff++ also improves sampling speed, reducing sampling time of FBSDiff from 8.0s to 6.1s.
	
	Ablation studies averaged over 20 samples w.r.t. the STP module for the extended application of style-specific content creation are reported in Tab. \ref{tab:stp_ablation}. The introduction of the STP module does not impair I2I style consistency (negligible difference in Style Distance) but slightly improves text fidelity (the gain in CLIP Similarity). Most importantly, the STP module substantially relaxes the I2I spatial correspondence constraint (cutting the Structure Similarity from 0.973 to 0.916), fulfilling the primary goal of arbitrary content manipulation without changing image style.

	\section{Conclusion}\label{sec13}
	This paper extends upon our previous conference paper, proposing FBSDiff++ which enhances our original method in efficiency, flexibility, and functionality. In efficiency, FBSDiff++ refines model architecture, saving 88.73\% inference time without compromising I2I quality. In flexibility, FBSDiff++ proposes AdaFBS which allows for input source images of arbitrary resolution and aspect ratio. In functionality, FBSDiff++ further enables localized image manipulation and style-specific content creation with only minor modifications to the core method. Our method is free from model training, finetuning, and optimization, all while exhibiting noticeable advantages in visual quality, versatility, and controllability. We hope our work provides novel insights about integration of generative AI with traditional signal processing and enlightens more frequency-based future research. 

	\bigskip

		
		
		
		
		
		
	\begin{appendices}
		
		\section{Preliminary background}\label{secA}
		\subsection{Diffusion model background}
		Since the advent of Denoising Diffusion Probabilistic Model (DDPM), diffusion model has soon dominated research field of generative AI due to its advantages in training stability and sampling diversity as compared with GAN. Grounded in the theory of stochastic differential equations, diffusion model learns to iteratively denoise a noise-corrupted input signal (e.g., an image or a video clip), ultimately generating clean data that follow the underlying target distribution. Diffusion model is conceptually composed of a forward diffusion process and a reverse denoising process. The forward diffusion process gradually adds noise to the data over a series of steps, transforming the data into a random Gaussian distribution, while the reverse denoising process learns to reverse the forward process by iteratively removing noise from the data, starting from pure noise and gradually reconstructing the original data. The model is trained to predict the noise added at each step of the forward process. By learning to denoise, the model can generate new data samples by starting from random noise and applying the reverse process.
		
		Given the original data distribution $q(x_{0})$, the forward diffusion process applies a $T$-step Markov chain to gradually add noise to the original data $x_{0}$ according to the conditional distribution $q(x_{t}|x_{t-1})$, which is defined as follows:
		\begin{equation}
			q(x_{t}|x_{t-1})=\mathcal{N}(x_{t}; \sqrt{\alpha_{t}}x_{t-1}, (1-\alpha_{t})\mathcal{I}),
		\end{equation}
		where $\alpha_{t}$ follows a predefined schedule, $\alpha_{t}\in(0, 1)$, $\alpha_{t} > \alpha_{t+1}$. Using the notation $\bar{\alpha}_{t}=\prod_{i=1}^{t}\alpha_{i}$, we can derive the marginal distribution $q(x_{t}|x_{0})$ that can be used to directly obtain $x_{t}$ from $x_{0}$ in a single step for arbitrary time step $t$:
		\begin{equation}
			q(x_{t}|x_{0})=\mathcal{N}(x_{t}; \sqrt{\bar{\alpha}_{t}}x_{0}, (1-\bar{\alpha}_{t})\mathcal{I}),
			\label{xt|x0}
		\end{equation}
		where $\sqrt{\bar{\alpha}_{T}} \approx 0$. With the forward diffusion process, the source data $x_{0}$ is transformed into $x_{T}$ that follows an isotropic Gaussian distribution.
		
		The reverse denoising process learns to conversely convert a Gaussian noise $x_{T}$ to the manifold of the original data distribution $q(x_{0})$ by gradually estimating and sampling from the posterior distribution $p(x_{t-1}|x_{t})$. Since the posterior distribution $p(x_{t-1}|x_{t})$ is mathematically intractable, we can derive the conditional posterior distribution $p(x_{t-1}|x_{t}, x_{0})$ with the Bayes formula and some algebraic manipulation:
		\begin{equation}
			p(x_{t-1}|x_{t},x_{0})=\mathcal{N}(x_{t-1}; \tilde{\mu}_{t}(x_{t}, x_{0}), \tilde{\beta}_{t}\mathcal{I}),
			\label{conditional_posterior}
		\end{equation}
		\begin{equation}
			\tilde{\mu}_{t}(x_{t}, x_{0})=\frac{\sqrt{\bar{\alpha}_{t-1}}\beta_{t}}{1-\bar{\alpha}_{t}}x_{0}+\frac{\sqrt{\alpha}_{t}(1-\bar{\alpha}_{t-1})}{1-\bar{\alpha}_{t}}x_{t},
		\end{equation}
		\begin{equation}
			\tilde{\beta}_{t}=\frac{1-\bar{\alpha}_{t-1}}{1-\bar{\alpha}_{t}}\beta_{t},
		\end{equation}
		where $\beta_{t}=1-\alpha_{t}$. 
		However, the conditional posterior distribution $p(x_{t-1}|x_{t}, x_{0})$ cannot be directly used for sampling since $x_{0}$ is unavailable at inference time ($x_{0}$ is the target of the sampling process). Thus, DDPM tries to estimate the unknown $x_{0}$ given the $x_{t}$ at each time step. Considering the reparameterization form of Eq. \ref{xt|x0}:
		\begin{equation}
			x_{t}=\sqrt{\bar{\alpha}_{t}}x_{0}+\sqrt{1-\bar{\alpha}_{t}}\epsilon_{t},
			\label{reparam}
		\end{equation}
		in which $\epsilon_{t}$ denotes the randomly sampled Gaussian noise that maps $x_{0}$ to $x_{t}$ in a single step according to Eq. \ref{xt|x0}. Given Eq. \ref{reparam}, we can represent $x_{0}$ using $x_{t}$ and $\epsilon_{t}$:
		\begin{equation}
			x_{0}=\frac{1}{\sqrt{\bar{\alpha}_{t}}}(x_{t}-\sqrt{1-\bar{\alpha}_{t}}\epsilon_{t}).
			\label{repre_x0_with_xt}
		\end{equation}
		
		However, the Gaussian noise $\epsilon_{t}$ sampled in the forward diffusion process is also unknown for the reverse denoising process where only $x_{t}$ is available. Consequently, DDPM builds a noise estimation network $\epsilon_{\theta}$ that predicts the sampled Gaussian noise $\epsilon_{t}$ in Eq. \ref{repre_x0_with_xt} with $x_{t}$ and time step $t$ as input, which is realized by training $\epsilon_{\theta}$ with the following noise regression loss:
		\begin{equation}
			L=\|\epsilon_{t}-\epsilon_{\theta}(x_{t}, t)\|_{2},
			\label{DDPM_loss}
		\end{equation}
		where $t\sim$ Uniform($\{1,...,T\}$), $\epsilon_{t} \sim \mathcal{N}(0, \mathcal{I})$, $x_{t}$ is computed via Eq. \ref{reparam}. After model training, $y_{\theta}(x_{t})$, the estimation of $x_{0}$ given $x_{t}$, can be obtained simply by replacing $\epsilon_{t}$ in Eq. \ref{repre_x0_with_xt} with the predicted noise $\epsilon_{\theta}(x_{t}, t)$:
		\begin{equation}
			y_{\theta}(x_{t})=\frac{1}{\sqrt{\bar{\alpha}_{t}}}(x_{t}-\sqrt{1-\bar{\alpha}_{t}}\epsilon_{\theta}(x_{t}, t)).
			\label{x_0_pred}
		\end{equation}
		
		Replacing the unknown $x_{0}$ in Eq. \ref{conditional_posterior} with its predicted estimation $y_{\theta}(x_{t})$ given by Eq. \ref{x_0_pred}, we can sample $x_{t-1}$ based on $x_{t}$ from the approximate posterior distribution $\mathcal{N}(x_{t-1};\tilde{\mu}_{t}(x_{t},y_{\theta}(x_{t})),\tilde{\beta}_{t}\mathcal{I})$, and thus sample the ultimate $x_{0}$ step by step from the initial Gaussian noise $x_{T}$. 
		
		\subsection{Conditional diffusion model}
		Taking the image generation task as an example, conditional diffusion model tackles conditional image synthesis by introducing additional condition $c$ to the model to guide image generation (denoising) process. In this paradigm, the condition signal $c$ together with $x_{t}$ and time step $t$ are taken as input to the noise estimation network $\epsilon_{\theta}$, such that $\epsilon_{\theta}$ is trained to conditionally predict the added Gaussian noise in the forward diffusion process, as supervised by the randomly sampled $\epsilon_{t}$ in Eq. \ref{reparam}. The training loss given by Eq. \ref{DDPM_loss} is correspondingly updated as:
		\begin{equation}
			L=\|\epsilon_{t}-\epsilon_{\theta}(x_{t}, t, c)\|_{2},
			\label{conditional_DDPM_loss}
		\end{equation}
		where $t\sim$ Uniform($\{1,...,T\}$), $\epsilon_{t} \sim \mathcal{N}(0, \mathcal{I})$, $x_{t}$ is computed via Eq. \ref{reparam}. After model training, the reverse sampling process is applied to generate new images from random Gaussian noise $x_{T}$, based on the step-by-step denoising according to the conditional posterior distribution given by Eq. \ref{conditional_posterior}, in which the unknown $x_{0}$ is approximated by the linear combination of $x_{t}$ and the conditional noise estimation, \textit{i.e.}, the $y_{\theta}(x_{t})$ (the approximate $x_{0}$ estimated by $x_{t}$) in Eq. \ref{x_0_pred} is updated as:
		\begin{equation}
			y_{\theta}(x_{t}, c)=\frac{1}{\sqrt{\bar{\alpha}_{t}}}(x_{t}-\sqrt{1-\bar{\alpha}_{t}}\epsilon_{\theta}(x_{t}, t, c)).
		\end{equation}
		
		\subsection{Denoising diffusion implicit model}
		Denoising diffusion implicit model (DDIM) is a variant of diffusion model that builds on the framework of DDPM but enables much more efficient sampling while maintaining high-quality generation results. DDIM can generate samples in significantly fewer steps compared with DDPM by modeling the reverse denoising process as a non-Markovian process and skipping the intermediate denoising steps. 
		
		DDIM is totally the same as DDPM in model training and only differs with DDPM in model inference, namely that DDIM can directly inherit the pre-trained DDPM model. To compute $x_{t-1}$ from $x_{t}$ in the reverse denoising (sampling) process, DDIM features a two-step deterministic denoising. In the first step, DDIM estimates an approximate $x_{0}$ based on $x_{t}$ using Eq. \ref{x_0_pred}. In the second step, DDIM computes $x_{t-1}$ from the approximate $x_{0}$ using the forward diffusion in the form of Eq. \ref{reparam}:
		\begin{equation}
			x_{t-1}=\sqrt{\bar{\alpha}_{t-1}}y_{\theta}(x_{t})+\sqrt{1-\bar{\alpha}_{t-1}}\epsilon_{t-1}, 
			\label{ddim_1}
		\end{equation}
		where $y_{\theta}(x_{t})$ is given by Eq. \ref{x_0_pred}. Considering that the $\epsilon_{t-1}$ in the above equation is the sampled Gaussian noise in the forward diffusion process, which is unknown in the reverse denoising process, we can replace $\epsilon_{t-1}$ with $\epsilon_{\theta}(x_{t-1}, t-1)$, the approximate $\epsilon_{t-1}$ estimated by the network $\epsilon_{\theta}$. Therefore, the Eq. \ref{ddim_1} can be updated as:
		\begin{equation}
			x_{t-1}=\sqrt{\bar{\alpha}_{t-1}}y_{\theta}(x_{t})+\sqrt{1-\bar{\alpha}_{t-1}}\epsilon_{\theta}(x_{t-1}, t-1). 
			\label{ddim_2}
		\end{equation}
		However, the $\epsilon_{\theta}(x_{t-1}, t-1)$ in the above equation is also unavailable since $x_{t-1}$ is unknown (we only know $x_{t}$ and want to compute $x_{t-1}$). Thus, we can further approximate $\epsilon_{\theta}(x_{t-1}, t-1)$ with $\epsilon_{\theta}(x_{t}, t)$ and arrive to the final DDIM sampling equation:
		\begin{equation}
			x_{t-1}=\sqrt{\bar{\alpha}_{t-1}}y_{\theta}(x_{t})+\sqrt{1-\bar{\alpha}_{t-1}}\epsilon_{\theta}(x_{t}, t). 
			\label{ddim_3}
		\end{equation}
		Eq. \ref{ddim_3} shows that the reverse sampling process of DDIM is totally deterministic, namely, each starting Gaussian noise $x_{T}$ yields a unique sampling result $x_{0}$. 
		
		Note that the above derived two-step sampling process of $x_{t} \rightarrow x_{0} \rightarrow x_{t-1}$ also applies for $x_{t} \rightarrow x_{0} \rightarrow x_{t+1}$. That is, a clean image $x_{0}$ can be deterministically inverted into a Gaussian noise through the following inversion process:
		\begin{equation}
			x_{t+1}=\sqrt{\bar{\alpha}_{t+1}}y_{\theta}(x_{t})+\sqrt{1-\bar{\alpha}_{t+1}}\epsilon_{\theta}(x_{t}, t). 
			\label{ddim_inversion}
		\end{equation}
		The DDIM inversion given by Eq. \ref{ddim_inversion} has wide applications in image editing and style transfer. For conditional image generation of DDIM, the $y_{\theta}(x_{t})$ and $\epsilon_{\theta}(x_{t}, t)$ in Eq. \ref{ddim_3} and Eq. \ref{ddim_inversion} are updated to $y_{\theta}(x_{t}, c)$ and $\epsilon_{\theta}(x_{t}, t, c)$ respectively.
		
		\subsection{Latent diffusion model}
		Latent diffusion model (LDM) compresses images from high-dimensional pixel space into low-dimensional feature space via pre-trained autoencoder, and builds diffusion model based on the latent feature space, such that computational overhead for both training and inference can be dramatically lowered. The training of LDM is similar to Eq. \ref{conditional_DDPM_loss} except that we use notation $z$ to denote latent features:
		\begin{equation}
			L=\|\epsilon_{t}-\epsilon_{\theta}(z_{t}, t, c)\|_{2},
			\label{LDM_loss}
		\end{equation}
		where $\epsilon_{t} \sim \mathcal{N}(0, \mathcal{I})$, $z_{t}=\sqrt{\bar{\alpha}_{t}}z_{0}+\sqrt{1-\bar{\alpha}_{t}}\epsilon_{t}$, $z_{0}=E(x_{0})$, $E$ is the pre-trained image encoder. The reverse denoising process from $z_{T} \sim \mathcal{N}(0, \mathcal{I})$ to $z_{0}$ is the same as $x_{T} \sim \mathcal{N}(0, \mathcal{I})$ to $x_{0}$ in DDPM. After reverse denoising process, the denoised clean features $z_{0}$ is decoded by the pre-trained decoder $D$ to yield the finally generated image $x_{0}$, \textit{i.e.}, $x_{0}=D(z_{0})$. In LDM framework, the condition $c$ could be either image features for I2I applications or textual features for T2I task.
		
		\section{User study}
\begin{figure*}[!htbp]
			\centering
			\includegraphics[width=0.9\linewidth]{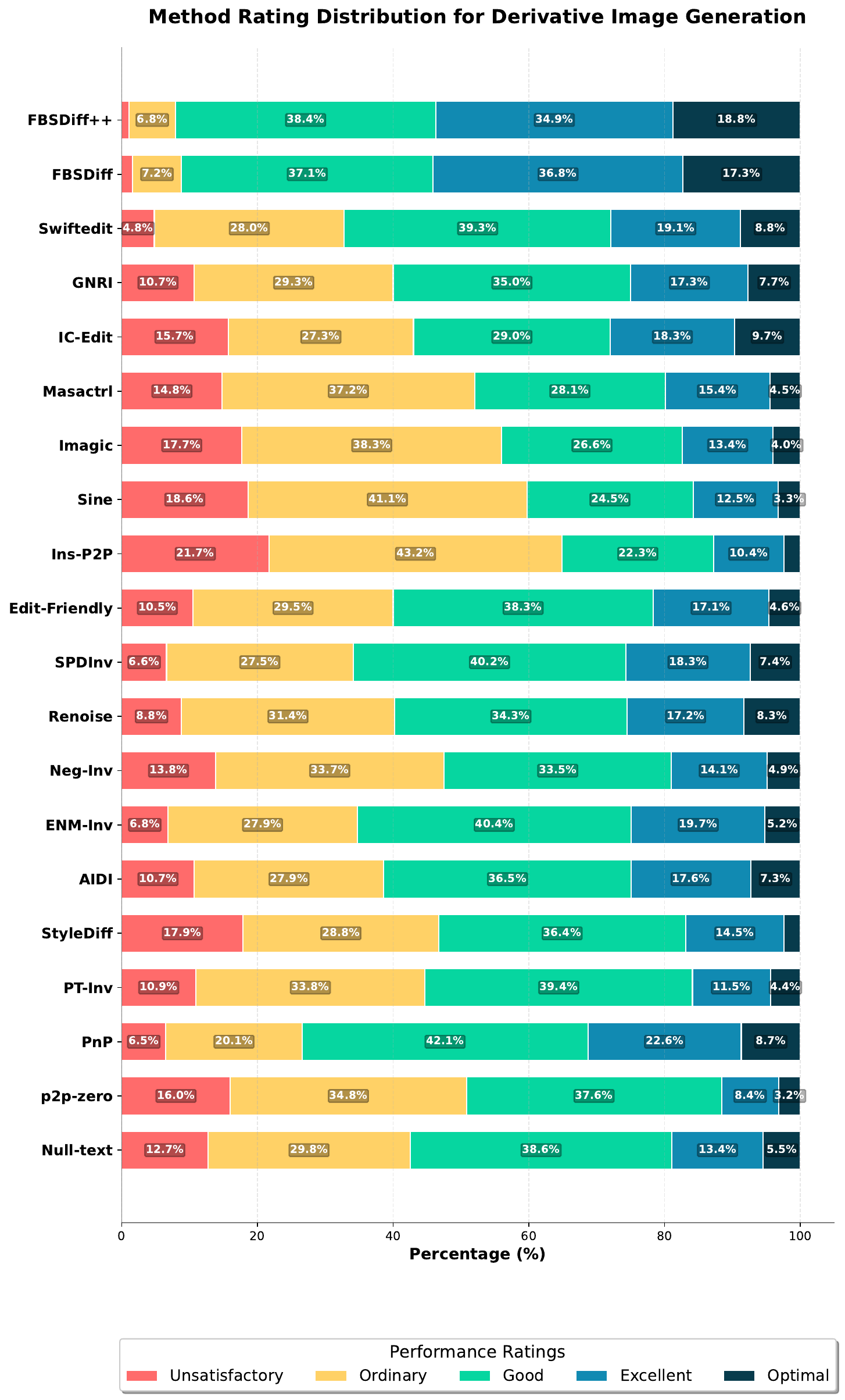}
			\caption{User study: method performance rating distribution for the I2I task of derivative image generation which favors I2I appearance consistency.}
			\label{fig:user_study_derivative_image_generation}
		\end{figure*}

\begin{figure*}[!htbp]
			\centering
			\includegraphics[width=0.9\linewidth]{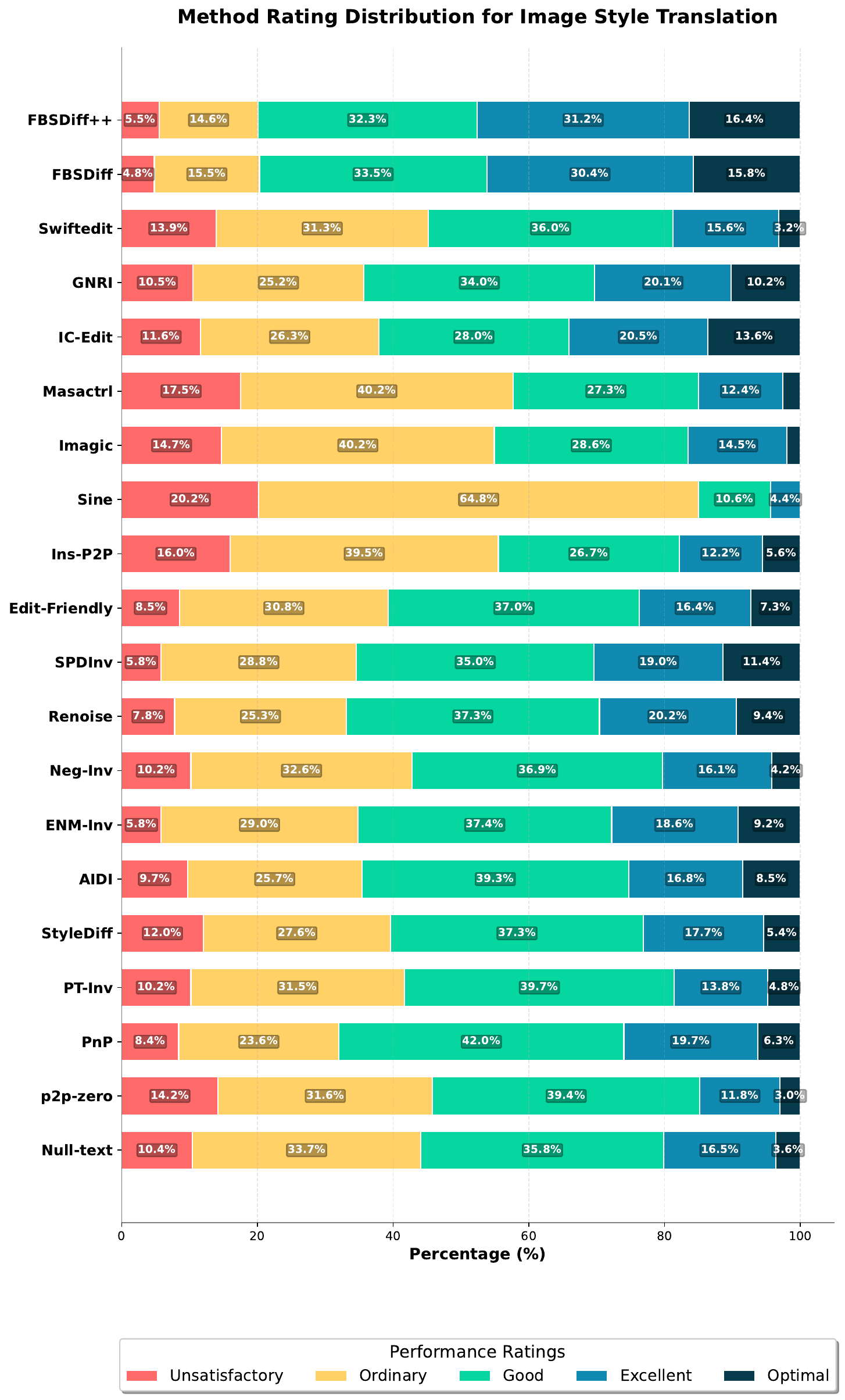}
			\caption{User study: method performance rating distribution for the I2I task of image style translation which favors I2I appearance divergence and diversity.}
			\label{fig:user_study_image_style_translatioin}
		\end{figure*}

		A user study is conducted to subjectively assess related methods. We poll 70 users to vote and rate results of 20 methods. The test samples include 200 pairs of source image and target text for the task of derivative image generation, and another 200 image-text pairs for the task of image style translation. For our proposed FBSDiff and FBSDiff++, we activate low-FBS mode for derivative image generation and high-FBS mode for image style translation. The rating scale is divided into five levels: unsatisfactory, ordinary, good, excellent, and optimal. We calculate the vote proportion of each rating level for each method. Statistical results shown in Fig. \ref{fig:user_study_derivative_image_generation} and Fig. \ref{fig:user_study_image_style_translatioin} demonstrate that FBSDiff and FBSDiff++ hold a significant advantage in the top two tiers for both two I2I tasks, indicating that our methods gain the most user preferences subjectively.
		
		\section{More algorithms}
		\begin{algorithm}[!htbp]
			\caption{Complete algorithm of FBSDiff++ for localized image manipulation}
			\label{FBSDiff++_algorithm}
			\begin{algorithmic}[1]
				\renewcommand{\algorithmicrequire}{\textbf{Input:}}
				\renewcommand{\algorithmicensure}{\textbf{Output:}}
				\REQUIRE{source image $x$, source image mask $M_{src}$, target text prompt.}
				\ENSURE{translated image $\tilde{x}$.}
				\STATE Extract the initial latent feature $z_{0}=E(x)$.
				\STATE Initialize an empty list $L$.
				\STATE Downsample $M_{src}$ to feature size $\rightarrow M_{f}$.
				\FOR{$t=0$ to $T-1$}
				\STATE compute $z_{t+1}$ from $z_{t}$ via Eq. \ref{eq:inversion};
				\STATE $L$.append($z_{t+1}$);
				\ENDFOR 
				\COMMENT{DDIM inversion}
				\STATE $L$.pop(-1). \COMMENT{remove the last element $z_{T}$}
				\STATE Initialize $\tilde{z}_{T} \sim \mathcal{N}(0, I)$.
				\FOR{$t=T$ to $\lambda T+1$}
				\STATE compute $\tilde{z}_{t-1}$ from $\tilde{z}_{t}$ via Eq. \ref{eq:sampling};
				\STATE $z_{t-1}$ = $L$.pop(-1);
				\STATE $\tilde{z}_{t-1}$ = AdaFBS($z_{t-1}$,$\tilde{z}_{t-1}$);
				\STATE $\tilde{z}_{t-1}$ = $\tilde{z}_{t-1}$ $\times$ $M_{f}$ + $z_{t-1}$ $\times$ ($1-M_{f}$);
				\ENDFOR\COMMENT{DDIM sampling w/ AdaFBS}
				\FOR{$t=\lambda T$ to $1$}
				\STATE compute $\tilde{z}_{t-1}$ from $\tilde{z}_{t}$ via Eq. \ref{eq:sampling};
				\STATE $\tilde{z}_{t-1}$ = $\tilde{z}_{t-1}$ $\times$ $M_{f}$ + $z_{t-1}$ $\times$ ($1-M_{f}$);
				\ENDFOR\COMMENT{DDIM sampling w/o AdaFBS}
				\STATE Obtain the final translated image $\tilde{x}=D(\tilde{z}_{0})$.
			\end{algorithmic}
		\end{algorithm}
		
		\begin{algorithm}[!htbp]
			\caption{Complete algorithm of FBSDiff++ for style-specific content creation}
			\label{FBSDiff++_algorithm}
			\begin{algorithmic}[1]
				\renewcommand{\algorithmicrequire}{\textbf{Input:}}
				\renewcommand{\algorithmicensure}{\textbf{Output:}}
				\REQUIRE{source image $x$, target text prompt.}
				\ENSURE{translated image $\tilde{x}$.}
				\STATE Extract the initial latent feature $z_{0}=E(x)$.
				\STATE Initialize an empty list $L$.
				\FOR{$t=0$ to $T-1$}
				\STATE compute $z_{t+1}$ from $z_{t}$ via Eq. \ref{eq:inversion};
				\STATE $L$.append($z_{t+1}$);
				\ENDFOR 
				\COMMENT{DDIM inversion}
				\STATE $L$.pop(-1). \COMMENT{remove the last element $z_{T}$}
				\STATE Initialize $\tilde{z}_{T} \sim \mathcal{N}(0, I)$.
				\FOR{$t=T$ to $\lambda T+1$}
				\STATE compute $\tilde{z}_{t-1}$ from $\tilde{z}_{t}$ via Eq. \ref{eq:sampling};
				\STATE $z_{t-1}$ = $L$.pop(-1);
				\STATE $z_{t-1}$ = STP($z_{t-1}$);
				\STATE $\tilde{z}_{t-1}$ = low-FBS($z_{t-1}$,$\tilde{z}_{t-1}$);
				\ENDFOR\COMMENT{DDIM sampling w/ AdaFBS}
				\FOR{$t=\lambda T$ to $1$}
				\STATE compute $\tilde{z}_{t-1}$ from $\tilde{z}_{t}$ via Eq. \ref{eq:sampling};
				\ENDFOR\COMMENT{DDIM sampling w/o AdaFBS}
				\STATE Obtain the final translated image $\tilde{x}=D(\tilde{z}_{0})$.
			\end{algorithmic}
		\end{algorithm}
		
		\begin{figure*}[!htbp]
			\centering
			\includegraphics[width=0.99\linewidth]{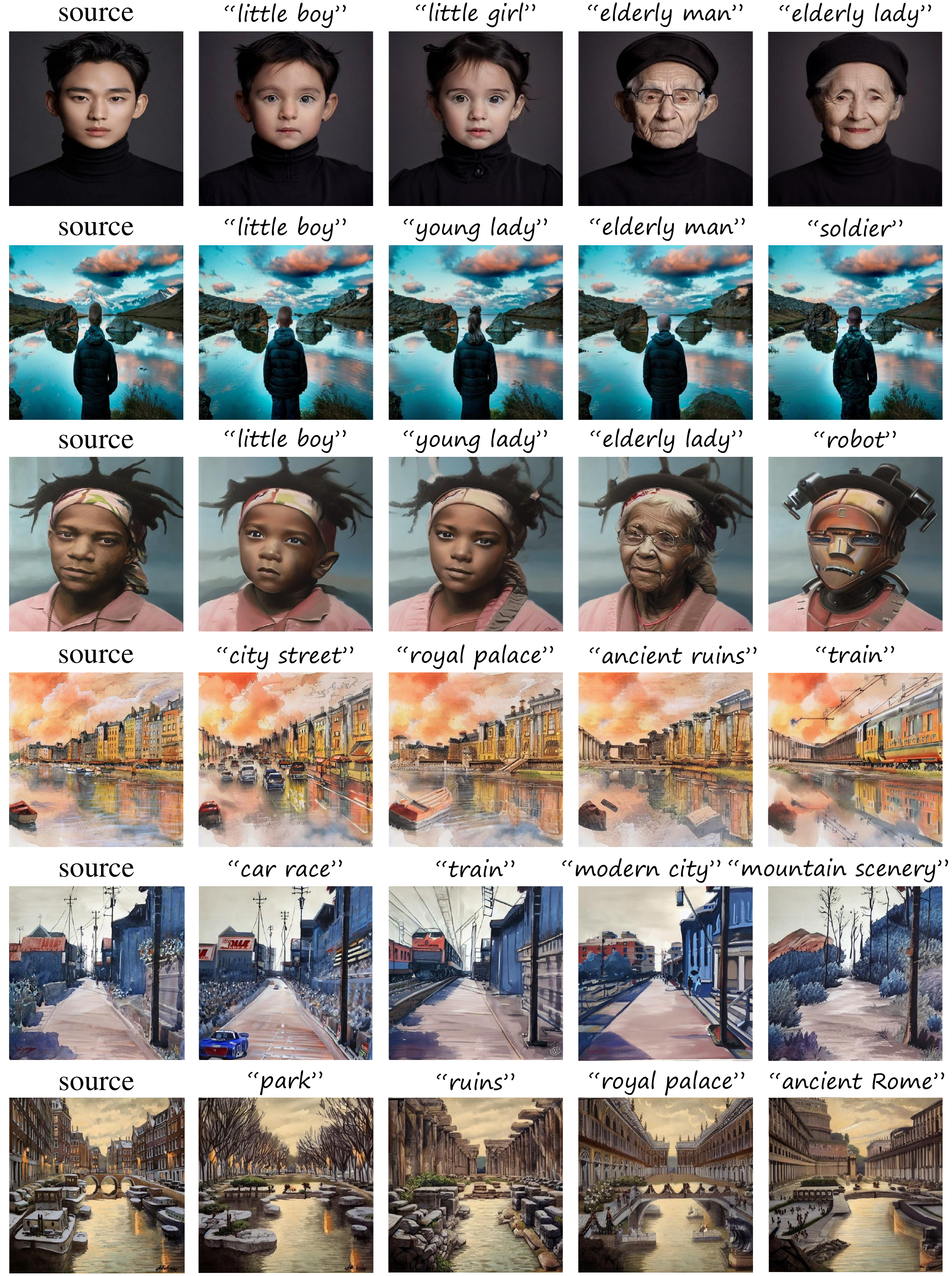}
			\caption{More text-driven I2I results of \textbf{FBSDiff} for derivative image generation.}
			\label{fig:FBSDiff_supp}
		\end{figure*}
		
		\begin{figure*}[!htbp]
			\centering
			\includegraphics[width=0.99\linewidth]{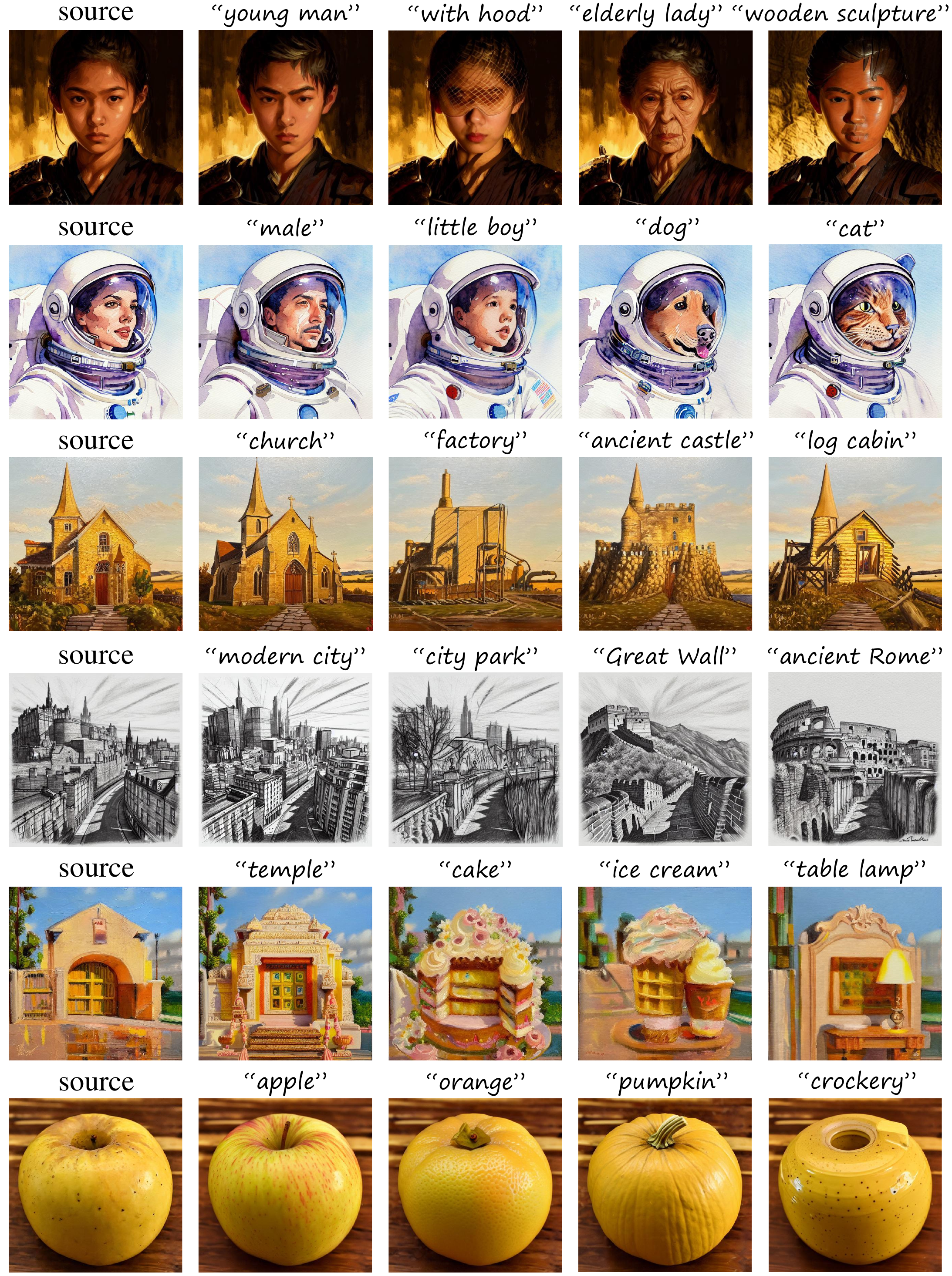}
			\caption{More text-driven I2I results of \textbf{FBSDiff++} for derivative image generation.}
			\label{fig:FBSDiff++_supp}
		\end{figure*}
		
		\begin{figure*}[!htbp]
			\centering
			\includegraphics[width=\linewidth]{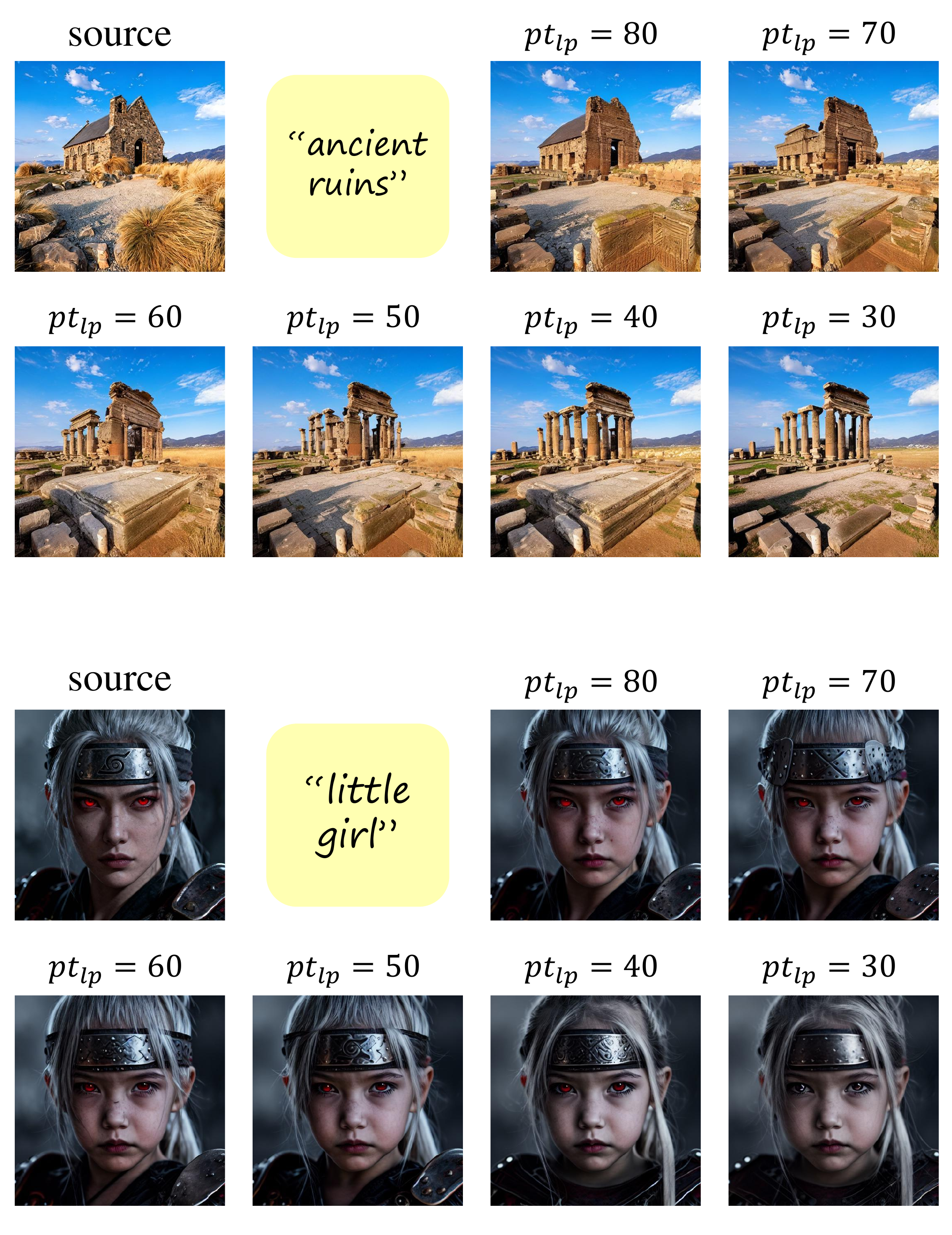}
			\caption{More I2I appearance correlation control of \textbf{FBSDiff++} achieved by tuning the low-FBS percentile threshold $pt_{lp}$.}
			\label{fig:FBSDiff++_control_supp}
		\end{figure*}
		
	\end{appendices}

	\bibliography{main}
	
\end{document}